\documentclass[10pt,twocolumn,letterpaper]{article}

\usepackage[pagenumbers]{cvpr} % To force page numbers, e.g. for an arXiv version

\usepackage{graphicx}
\usepackage{amsmath}
\usepackage{amssymb}
\usepackage{booktabs}
\usepackage{enumitem}
\setlist[itemize]{noitemsep}
\setlist{nosep}

\usepackage[pagebackref,breaklinks,colorlinks]{hyperref}

\usepackage[capitalize]{cleveref}
\crefname{section}{Sec.}{Secs.}
\Crefname{section}{Section}{Sections}
\Crefname{table}{Table}{Tables}
\crefname{table}{Tab.}{Tabs.}

\usepackage{enumitem,amssymb}
\newlist{todolist}{itemize}{2}
\setlist[todolist]{label=$\square$}
\usepackage{pifont}
\def\cvprPaperID{5675} % *** Enter the CVPR Paper ID here
\def\confName{CVPR}
\def\confYear{2025}

\begin{document}

\title{FathomVerse: A community science dataset for ocean animal discovery
\\ \vspace{2mm} \includegraphics[width=1\textwidth]{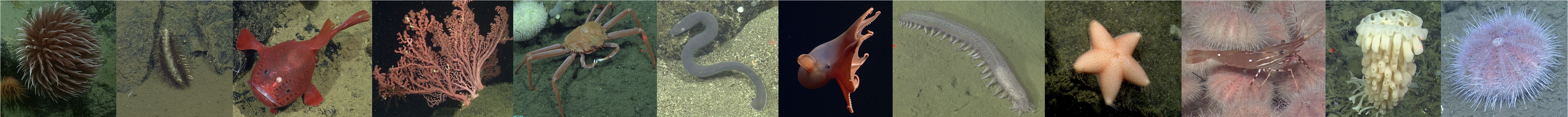}}

%%% Authors: Genevieve Patterson, Joost Daniels, Benjamin Woodward, Kevin Barnard, Giovanna Sainz, Lonny Lundsten, Kakani Young

\author{
\textbf{Genevi\`{e}ve Patterson\textsuperscript{2} \hfill Joost Daniels\textsuperscript{1} \hfill Benjamin Woodward\textsuperscript{3}}\\
\textbf{Kevin Barnard\textsuperscript{1} \hfill Giovanna Sainz\textsuperscript{1} \hfill Lonny Lundsten\textsuperscript{1} \hfill Kakani Katija\textsuperscript{1}}\\
\textsuperscript{1}Monterrey Bay Aquarium Research Institute (MBARI) \\  \textsuperscript{2}Barderry Applied Research \hfill  \textsuperscript{3}CVision AI
}
% \and
\maketitle

%%%%%%%%% ABSTRACT
\begin{abstract}
Can computer vision help us explore the ocean? The ultimate challenge for computer vision is to recognize any visual phenomena, more than only the objects and animals humans encounter in their terrestrial lives. Previous datasets have explored everyday objects and fine-grained categories humans see frequently. We present the FathomVerse v0 detection dataset to push the limits of our field by exploring animals that rarely come in contact with people in the deep sea. These animals present a novel vision challenge. 

The FathomVerse v0 dataset consists of 3843 images with 8092 bounding boxes from 12 distinct morphological groups recorded at two locations on the deep seafloor that are new to computer vision. It features visually perplexing scenarios such as an octopus intertwined with a sea star, and confounding categories like vampire squids and sea spiders. This dataset can push forward research on topics like fine-grained transfer learning, novel category discovery, species distribution modeling, and carbon cycle analysis, all of which are important to the care and husbandry of our planet.
\end{abstract}

%%%%%%%%% BODY TEXT

\section{Introduction }%- KAKANI}
\label{sec:intro}
The world ocean is largely unexplored and is the largest habitable ecosystem on our planet\cite{webb2010biodiversity}. Only 7\% of the ocean's surface is covered by long-term biological observations\cite{appeltans2012magnitude}, and given the ocean's volume, there are massive gaps to fill. Because of our lack of exploration and capacity, it is estimated that anywhere from 30 to 60\% of marine life is unknown to science\cite{webb2010biodiversity}. 

Biological observations include approaches that involve sampling environmental DNA, sound, and imagery, however imaging is the most direct, non-invasive method. Visual data are collected from a number of platforms to identify, quantify, and characterize the biological communities found throughout the ocean. This underwater world is visually distinct from the terrestrial environment, and ocean animal recognition and discovery is a huge challenge for transfer learning and could be a compelling test bench for novel research in category discovery\cite{han2020automatically,han2021autonovel}.

The FathomNet Database \cite{katija2022fathomnet} aggregates labeled imagery from marine researchers around the world. Annotating these images can be time-consuming and expensive, taking up the precious hours of expert marine biologists and taxonomists. While FathomNet can be used to aggregate labeled data, generation of labeled data is prohibitive for experts, and is desparately needed for training models to generalize to a variety of geolocations and concepts. We want to scale the annotation of all ocean animals (found throughout benthic, midwater, and shallow water habitats) for the conservation and expansion of scientific knowledge of those species given the urgent pressures these communities face with climate change and growing human activities. 

To that end, we designed and built the FathomVerse game, a community science effort to collect consensus annotations from ocean enthusiasts and game players anywhere in the world to images contributed to the FathomNet Program. Players are presented with regions of interest and are asked to classify those images into 12 distinct morphological groups. This paper presents the first expert-reviewed dataset created by the FathomVerse game -- FathomVerse v0 -- that was generated during beta testing in summer of 2023. In the following sections we describe:
\begin{itemize}
\item The FathomVerse Game Design
\item Player Training
\item Game Release Schedule and Player Testing
\item Player Annotation Analysis and Trust Building
\item FathomVerse v0 Dataset Construction
\item Detector Experiments -- Successes and Failures
\end{itemize}

FathomVerse is now available on the iOS and Google Play App stores\cite{ios_app_store,google_play_store}.

\section{Related Work}% - GEN}
\label{sec:related_work}
Marine species are vastly understudied \cite{mora2011many}, and deep sea habitats in particular are chronically under-explored \cite{webb2010biodiversity}. We believe that games with a purpose \cite{von2006games} could be a key to helping understand the secrets of the deep on a global scale, by employing participatory science through engaging, science-based apps.

Other available scientific projects, such as the protein-folding game FoldIt \cite{cooper2010predicting}, have had significant impact by employing internet-scale participation by science enthusiasts. Animal conservation benefits immensely from computer vision dataset collection efforts such as the Caltech UCSD Birds (CUB-200) Dataset\cite{wah2011caltech}, NABirds app\cite{van2015building}, eBird app\cite{sullivan2014ebird}, Merlin Photo ID App \cite{van2015building}, and iNaturalist Dataset \cite{van2018inaturalist} to name some of the most popular apps and datasets. Although not tied to a particular science objective, the obsession of casual game players to answer exacting technical questions has led to the web-based GeoGuessr game, where players guess the GPS coordinates of an unlabeled street view from anywhere in the world, to become the one of the most popular browser games in the world\cite{geoguessr}. Combining gaming with a purpose with the mission of ocean life discovery provides an opportunity for broad community engagement.

Within the marine science community, the FathomNet\cite{boulais2020fathomnet,katija2022fathomnet} program, along with efforts to understand the needs of marine scientists with limited machine learning experience\cite{crosby2023designing}, has led to the development of the FathomNet Database API and dataset\cite{katija2022fathomnet}, subsequent challenges such as the FathomNet Challenges 2023 and 2024\cite{orenstein2023fathomnet2023}, and the FathomGPT API \cite{khanal2024fathomgpt}.

Here we present a dataset derived from a game inspired by the machine-teaching literature \cite{johns2015becoming} and human-in-the-loop systems \cite{patterson2015tropel} that allow people to directly contribute to ocean science. We borrow from projects that pioneered fine-grained recognition for conservation \cite{beery2023wild,beerymegadetector,tuia2022perspectives} and machine learning for images from nature \cite{Van_Horn_2021_CVPR}. Ultimately, we hope that our efforts will contribute to the managed mitigation and adaptation of the world to climate change\cite{rolnick2022tackling,cardinale2012biodiversity}.

\section{Game Design and Player Education}% - KAKANI}
\begin{figure}[h!]
    \centering
    \begin{subfigure}[t]{0.4\linewidth}
        \centering
        \includegraphics[width=0.75\linewidth]{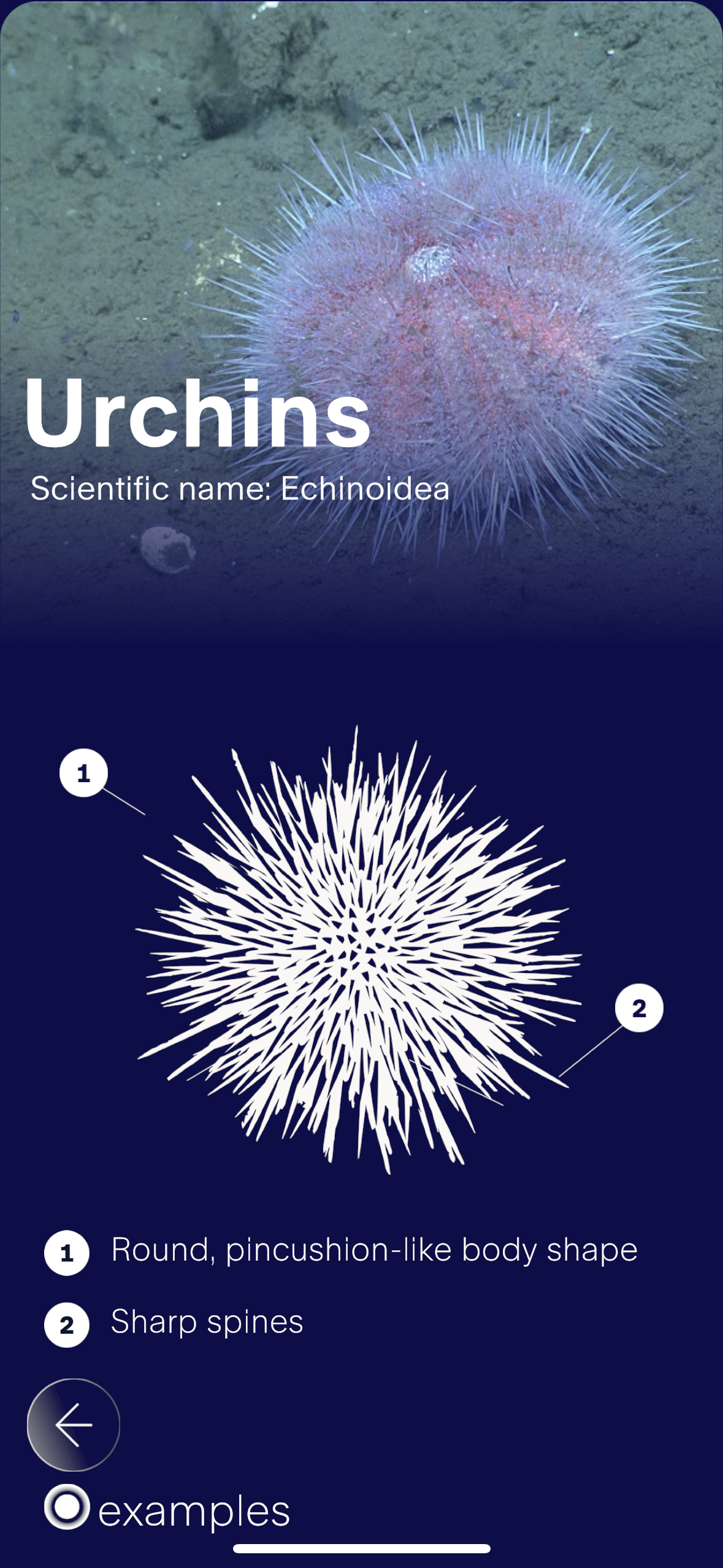}
        \caption{Active Missions}
        %\caption{Active Missions - Players `dive' to collect three mission animals, and learn about these animals in Mission Briefings.}
        \label{fig:mission_briefing}
    \end{subfigure}%
    \hspace{1em}
    \begin{subfigure}[t]{0.4\linewidth}
        \centering
        \includegraphics[width=0.75\linewidth]{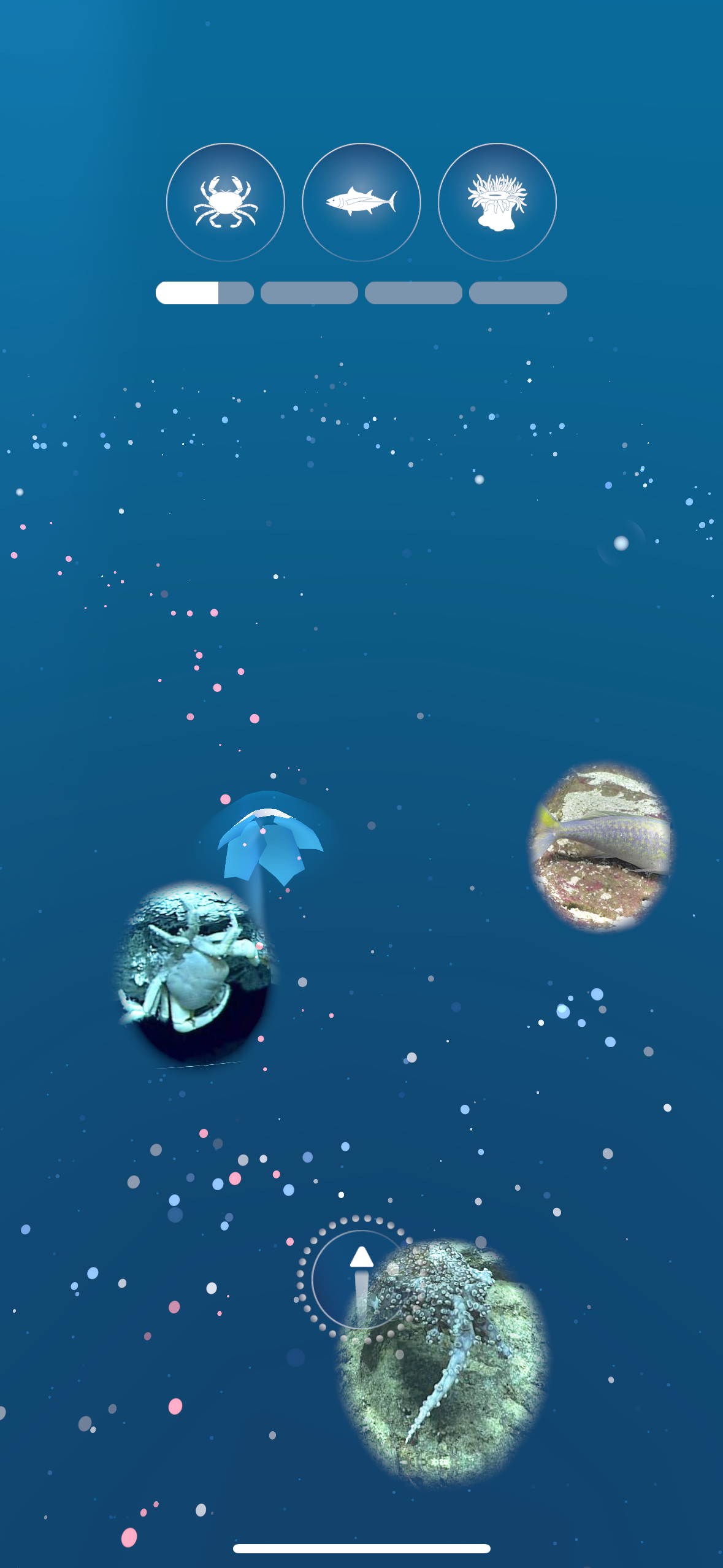}
        \caption{Collection}
        %\caption{Collection - Players experience the deep via an avatar that swims with the currents. Animals appear in bubbles that can be collected.}
        \label{fig:game_currents}
    \end{subfigure}%
    \\
    \begin{subfigure}[t]{0.4\linewidth}
        \centering
        % \includegraphics[width=0.75\linewidth]{figures/game_log_view.png}
        % \caption{Player Log - Animals are collected into a Player's Log.}
        \includegraphics[width=0.75\linewidth]{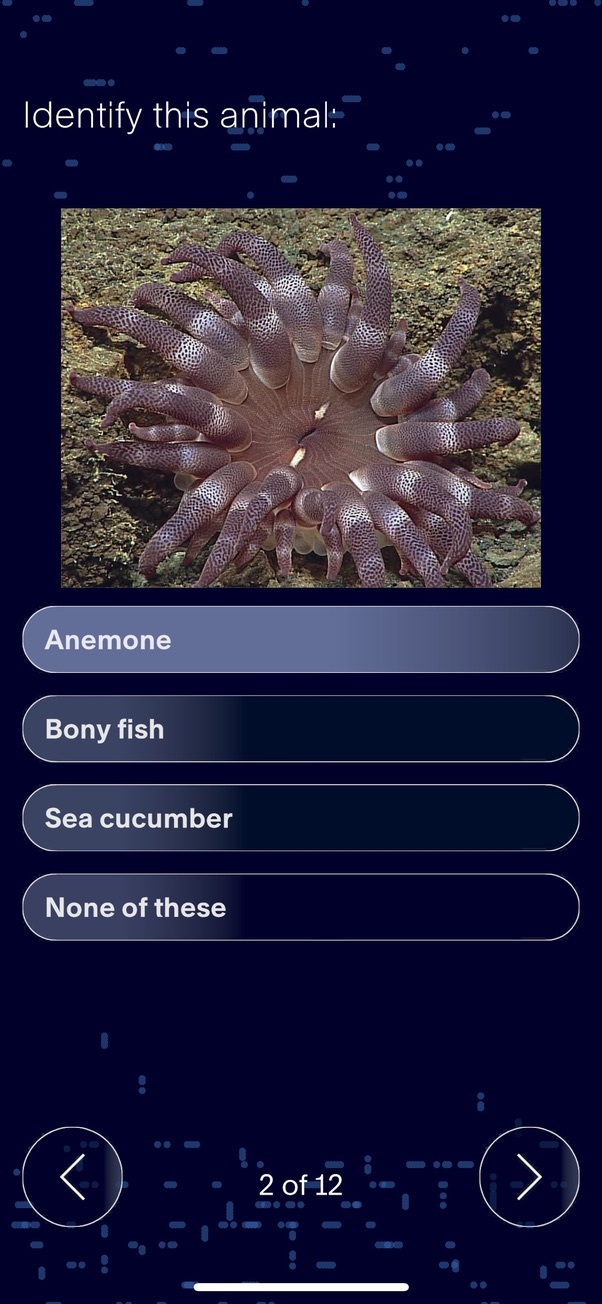}
        \caption{Labeling}
        %\caption{Labeling - Players select one of the three mission animal icons or \textit{None of the These} to label each animal they collected while diving.}
        \label{fig:game_annotation}
    \end{subfigure}%
    \hspace{1em}
    \begin{subfigure}[t]{0.4\linewidth}
        \centering
        \includegraphics[width=0.75\linewidth]{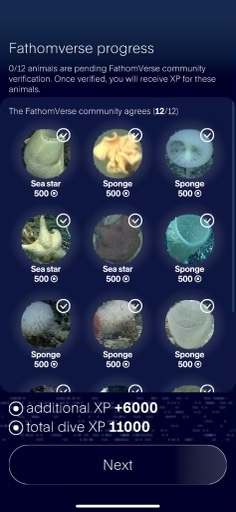}
        \caption{Verification}
        %\caption{Verification - After players submit their annotations, feedback (e.g., correct, incorrect, missed) is shown to the player.}
        \label{fig:game_score}
    \end{subfigure}%
    \caption{Various screens in FathomVerse v0 Game. (a) Active Missions - Players `dive' to collect three mission animals, and learn about these animals in Mission Briefings. (b) Collection - Players experience the deep via an avatar that swims with the currents. Animals appear in bubbles that can be collected. (c) Labeling - Players select one of the three mission animal icons or \textit{None of the These} to label each animal they collected while diving. (d) Verification - After players submit their annotations, feedback (e.g., correct, incorrect, missed) is shown to the player.}
    \label{fig:game-views}
\end{figure}

\label{sec:game}
After downloading the FathomVerse game to their phone or tablet, players can select twelve different animals -- anemone, bony fish, coral, crab, eel-like fish, octopus, sea cucumber, shrimp, sponge, urchin, and benthic worm. These animal groups serve as missions in the game, where players are looking for imagery that contain their active mission. Mission briefings [Fig.~\ref{fig:game-views}(a)] provide additional context around specific animal missions, and contain imagery of representative (positives) and confounding animals (hard negatives), as well as an illustration indicating morphological features that are used to distinguish the animal group from others.  We do not dynamically show players optimally selected educational images such as in Singla et al.\cite{singla2014near}.

\begin{figure}[h!]
    \centering
    \includegraphics[width=\linewidth]{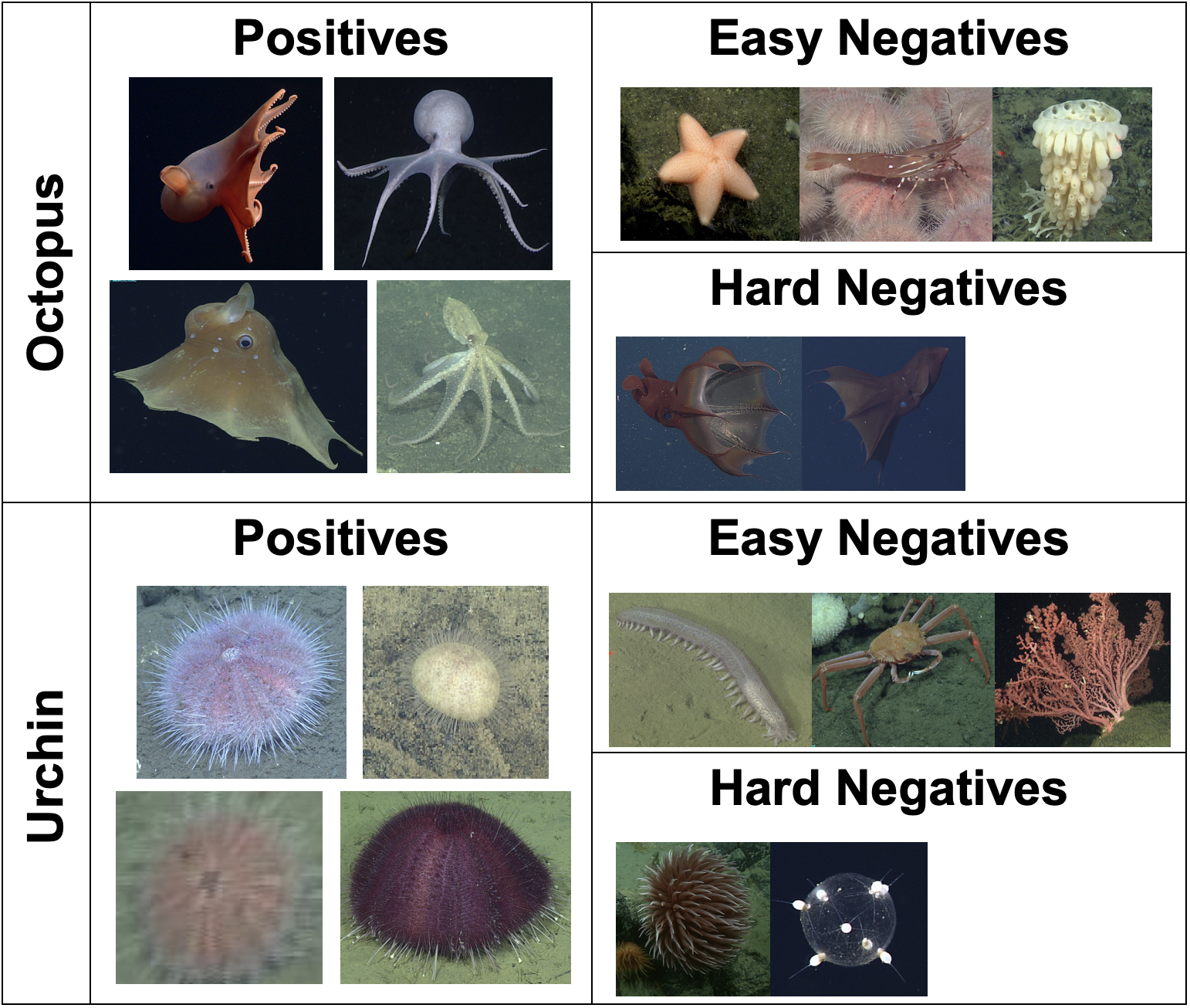}
    \caption{Dive imagery. For each active animal mission group, players are shown images that are either positives, easy negatives, or hard negatives. Example images for octopus (top row) and urchin (bottom row) are shown. Hard negatives for octopus include vampire squid; hard negatives for urchin include anemones and radiolarians.}
    \label{fig:hard-neg-ex}
\end{figure}

After selecting up to three animal missions, players can go on a `dive' [Fig.~\ref{fig:game-views}(b)], where they look for the active mission animals (indicated by the three animal icons located at the top of the screen) among 25-50 randomly selected images from a repository containing expert-labeled images \cite{katija2022fathomnet} and unlabeled images collected by a deep-diving remotely operated vehicle. These images were captured by Remotely operated vehicles (ROVs) exploring the ocean floor, capturing imagery likely seen only by the scientific team. Images shown in the dive include either positive, easy negative, or hard negative images for each active animal mission [Fig.~\ref{fig:hard-neg-ex}].

Players collect animals until they tire or see all of the animals available in the dive. Then they are guided to an annotation screen [Fig.~\ref{fig:game-views}(c)], where the player can select a label for the given ROI image. After the player has annotated all of their collected animals, they learn which expert-labeled animals the annotated correctly and incorrectly [Fig.~\ref{fig:game-views}(d)]. Players are rewarded with game points (or contribution points) for collecting animals during the dive and for labeling images correctly. Players increase their player Level (1-15) as they earn contribution points. 

\section{Game Experiments}% - KAKANI (Wave Results)}
\label{sec:game_waves}
FathomVerse v0 game was released to a select group of beta testers who signed up to participate in the trial several months or several days before they could download the app. To evaluate the annotation potential of players sourced from different communities, as well as the difficulty of annotating images sourced from different locations and imaging systems, we split up the game release into four waves (A-D), each lasting about two weeks. Players downloaded the game via Apple's TestFlight or Google's Play Console testing platforms using a link that was shared at specific times during the testing phase. Players that joined in an earlier wave could continue playing as long as the test app was available. 

\begin{figure*}[h!]
    \centering
    % \begin{subfigure}[t]{0.45\textwidth}
    %     \centering
    %     \includegraphics[width=0.9\linewidth]{figures/raw_player_annotations_per_wave.png}
    %     \caption{ }%Number of annotations per animal mission group across waves A-D.}
    %     %\caption{Differences in participation from each wave. Waves A-C, even though they included players of different demographics and inclinations, had similar participation. Wave D had more overall players and thus more contributed annotations.}% }
    %     \label{fig:wave_annotations}
    % \end{subfigure}%
    % ~
    % \begin{subfigure}[t]{0.45\textwidth}
    %     \centering
        \includegraphics[width=\linewidth]{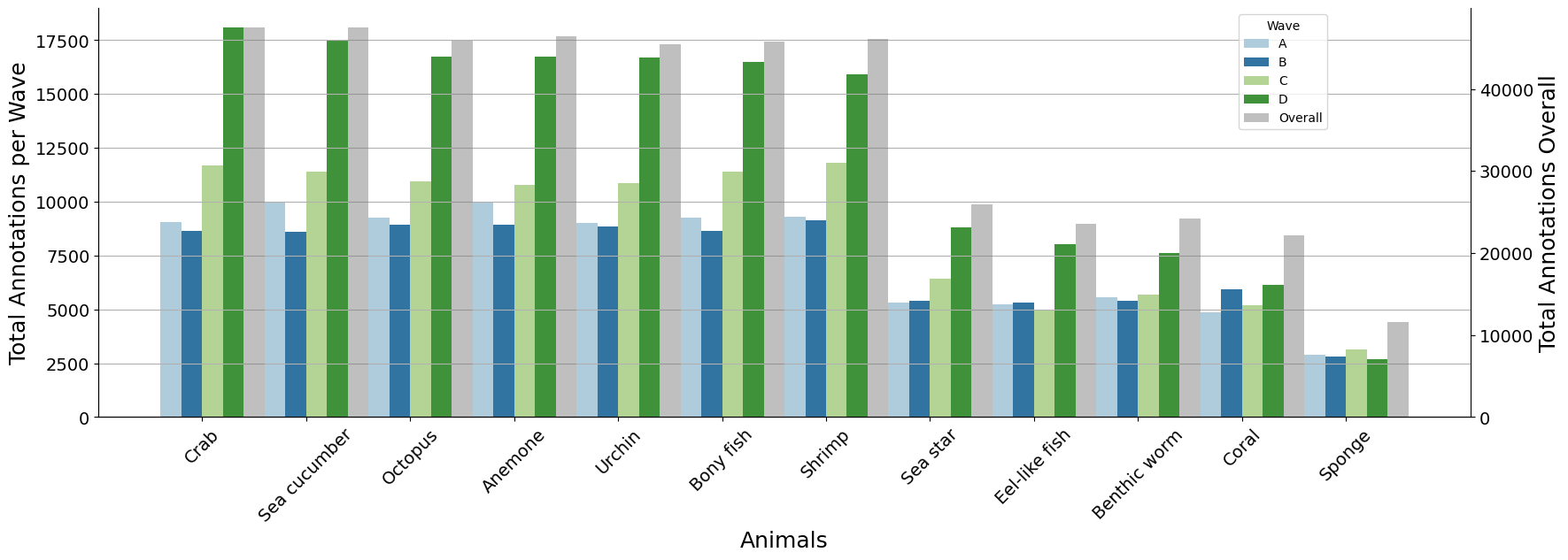}
    %     \caption{ }%Overall number of annotations during beta testing per animal mission group.}
    %     %\caption{Cumulative annotations for each animal across all waves.}% }
    %     \label{fig:overall_annotations}
    % \end{subfigure}
    \caption{Raw player annotation counts during beta testing. Left axis tracks number of annotations per animal mission group across waves A-D. Right axis tracks overall number of annotations during beta testing per animal mission group.}
    \label{fig:player_annotation_counts}
\end{figure*}

Waves A and B were open to people who signed up on the FathomVerse website. These players found out about FathomVerse from social media and ocean enthusiast communities months to weeks in advance, where ocean enthusiasts tend to have a cursory knowledge of ocean life. Waves C and D were opened to the general public. Waves A and B were shown images from the Octopus Garden, and waves C and D used imagery from the Musicians Seamounts; image sources will be described more in detail later. Statistics on total number of annotations across each wave and animal mission group is shown in Table~\ref{tab:player_counts} and Fig.~\ref{fig:player_annotation_counts}.

In Fig. \ref{fig:player_annotation_counts} it is notable that the Sponge class received similarly few annotations across waves, highlighting a need for more aggressive collection of Sponge annotations in future iterations of FathomVerse data collection. Figure \ref{fig:player_annotation_counts} reveals that half of the classes received between 45,000-50,000 annotations, while the other half of the classes received about half as many annotations. This does not reflect the expected natural animal populations in the benthic environments surveyed. In future iterations of data collections we will be challenged to improve the initial objectness detector that suggests bounding boxes for annotations as well as player eduction. The lower-annotated classes are often difficult to annotate (except sea star), even for educated marine biologists, due to the visual similarity of those animals and the substrate material they live on/in.

\begin{table}[h!]
\caption{Statistics covering per Player Annotation Capacity in Each Wave}
\resizebox{0.5\textwidth}{!}{%
\begin{tabular}{lrrrrrrrr}
\toprule
 & \# Players & Mean \#Anns & Std Anns & Min Anns & 25\% & 50\% & 75\% & Max Anns\\
Wave &  &  &  &  &  &  &  &  \\
\midrule
A & 215 & 318.88 & 450.41 & 3 & 28 & 133 & 443 & 2885 \\
B & 184 & 186.07 & 307.49 & 2 & 20 & 77 & 235.25 & 2934 \\
C & 188 & 62.58 & 86.90 & 1 & 6 & 21.50 & 76.50 & 521 \\
D & 276 & 43.22 & 52.91 & 0 & 4 & 25.50 & 55.25 & 280 \\
Total & 863 & 146.57 & 292.69 & 0 & 11 & 43 & 149 & 2934 \\
\bottomrule
\end{tabular}
}
\label{tab:player_counts}
\end{table}

\section{Player Annotation Analysis}% - GEN}
\label{sec:player_analysis}
 The four release waves gave us useful comparisons between users of varying communities. This section discusses how we evaluated the contributions of individual players, which players contributed to the FathomVerse v0 dataset, and what improvements to player education we should make based on our findings. 

To review the unlabeled images shown to users during Waves A-D, the first two waves showed images from the Musicians Seamounts, which are located in the Pacific Ocean (just north of the Hawaiian Islands) with an average depth 5,200 m (17k ft). The images were recorded by the ROV \textit{Deep Discoverer}; here we used the full HD resolution (1920x1080 px) science camera footage. The second two waves showed images from the Octopus Garden, which is located 3,200 m (10.5k ft) deep near the base of Davidson Seamount, an inactive underwater volcano 130 kilometers (80 miles) southwest of Monterey, California.  Imagery from the Octopus Garden were collected using the main science camera on the ROV \textit{Doc Ricketts} deployed from the R/V \textit{Western Flyer}. This camera was a prototype version of the MxD Seacam (DeepSea, San Diego, CA, USA) developed by DeepSea in collaboration with the authors' institution, recording imagery in 4K UHD (3840 x 2160 px). 

We use annotations from a marine biologist among the authors of this work for all bounding boxes from the Octopus Garden (9,919 positive and negative labels). Bounding boxed regions of interest (ROIs) were suggested by an objectness pre-filter and the expert classified those ROIs. Annotations for \textit{urchin} and \textit{anemone} were sparse in the Octopus Garden, and so we use NOAA expert annotations from the Musicians Seamount images for those categories (1,211 positive and negative labels). ROIs shown to players were selected from the raw unlabeled images via a YOLOv8 detector\cite{varghese2024yolov8}, trained on the full FathomNet dataset training set\footnote{\url{https://huggingface.co/FathomNet/megalodon}}. Any bounding boxes with confidence \textgreater 0.1, sparsified via NMS, were sent to players for annotation. We considered this a weak objectness pre-filter, but later results in this paper will show that this pipeline choice may have limited our ability to annotate animals that blend well with the background. 

\begin{figure}[h!]
    \begin{subfigure}[t]{0.45\linewidth}
        \centering
        \includegraphics[width=\linewidth]{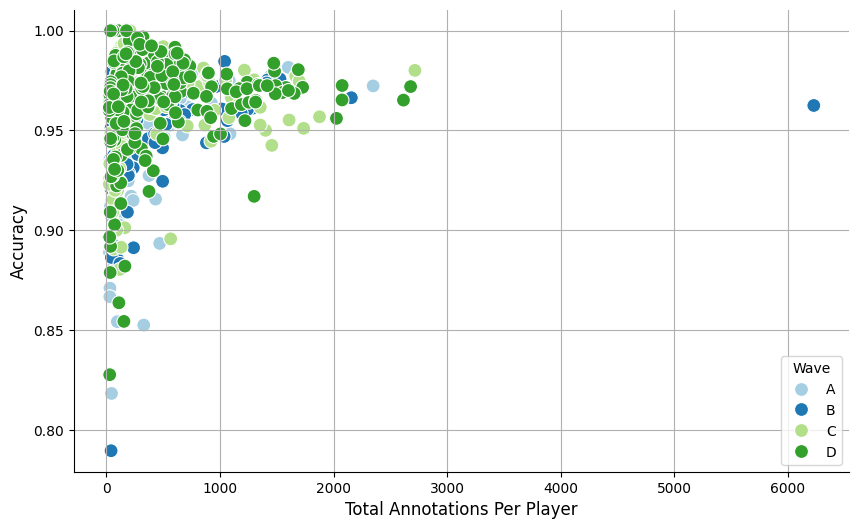}
        \caption{ }%The above scatter plot plots the accuracy of each player compared to labels from a marine biologist familiar with benthic animals. The dot colors show which wave a player belonged in. The player accuracies are mostly overlapping, with a slight increase in number of annotations and accuracy for players that joined during the first two waves, which were targeted at the ocean enthusiast audience. The overall highest contributing player is from Wave B, and is likely an ocean enthusiast.}
        \label{fig:player_acc}
    \end{subfigure}
    ~
    % \begin{subfigure}[t]{0.5\textwidth}
    %     \centering
    %     \includegraphics[height=1.2in]{figures/prec_vs_total_anns_per_player_expert_giovanna_fn_urchin_anemone.png}
    %     \caption{Lorem ipsum, lorem ipsum,Lorem ipsum, lorem ipsum,Lorem ipsum}
    % \end{subfigure}
    % \begin{subfigure}[t]{0.5\textwidth}
    %     \centering
    %     \includegraphics[height=1.2in]{figures/rec_vs_total_anns_per_player_expert_giovanna_fn_urchin_anemone.png}
    %     \caption{Lorem ipsum, lorem ipsum,Lorem ipsum, lorem ipsum,Lorem ipsum}
    % \end{subfigure}
    \begin{subfigure}[t]{0.45\linewidth}
        \centering
        \includegraphics[width=\linewidth]{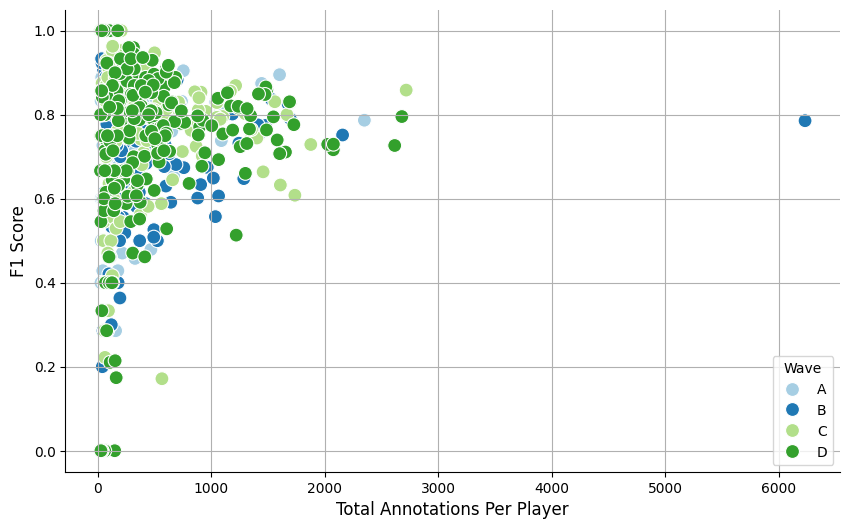}
        \caption{ }%The above plot plots Player F1 scores, which have a similar trend to player accuracy, indicating that players had reasonable Recall, finding as many animals as they could and not only selecting the easiest to recognize animals.}
        \label{fig:player_f1}
    \end{subfigure}
    \caption{Individual player performance. (a) Accuracy of each player compared to labels from an expert marine biologist. (b) Player F1 scores compared to labels from an expert. Colors indicate players from different release waves of the game experiment: blue dots are for the waves labeling data from Musicians Seamount; green dots are for the waves labeling Octopus garden images. The overall highest contributing player is from Wave B, and is likely an ocean enthusiast.}
    \label{fig:player_metrics}
\end{figure}

Player accuracy and F1 score (the harmonic mean of precision and recall) were generally high in the FathomVerse v0 dataset (Fig.~\ref{fig:player_metrics}), however the F1 scores varied greatly. It is worth reiterating that players joined during a given wave, but continued playing in later waves. The highest contributor, by an enormous margin, joined during Wave B, marking them as an Ocean Enthusiast who likely follows the livestreams of research dives. Wave D, which was open to the general public, had the most players. There does not appear to be a qualitative difference between the players from the different waves, with the exception of the stand-out, \#1 contributor.

\section{FathomVerse v0 Datatset}% - GEN}
\label{sec:dataset}
If marine biologists with graduate degrees and taxonomic specialization could annotate every pixel in an image dataset, theoretically we would have a high quality dataset. However, that approach is not scalable as achieving a global dataset requires large-scale effort that exceeds the number of individuals with that expert knowledge. Here, we strive to create a high-quality dataset using FathomVerse player annotations with the following features: highest precision and recall, the most correct labels, and the largest range of visual appearance for the dataset concepts.  

\begin{figure}[h!]
    \centering
    \begin{subfigure}[t]{0.45\linewidth}
        \centering
        \includegraphics[width=\linewidth]{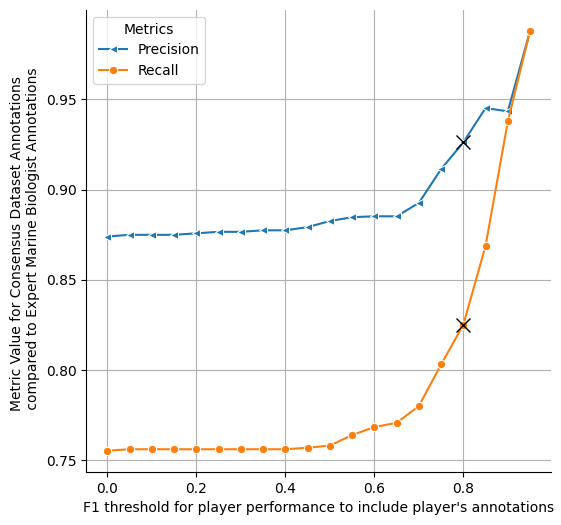}
        \caption{ }%The authors wanted to construct a dataset of consensus annotation with the highest precision and recall (as compared to experts) as possible. The above trend lines show how the precision and recall of the consensus dataset would vary if only annotations from players with F1 scores higher than the x-axis threshold were included. For FathomVerse v0, the authors elected to use a player F1 threshold of 0.8, marked with a black X on the above graph.}
        \label{fig:player_threshold_f1}
    \end{subfigure}%
    ~
    \begin{subfigure}[t]{0.45\linewidth}
        \centering
        \includegraphics[width=\linewidth]{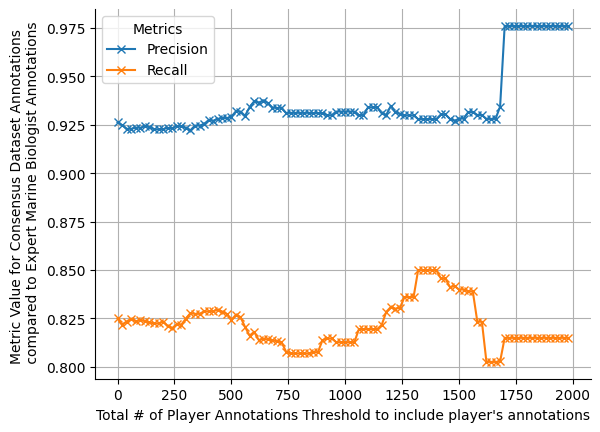}
        \caption{ }%The above plot shows how thresholding players by their total number of annotations, given they achieved at least an F1 score of 0.8 (see previous plot), would affect the quality of the Consensus Dataset. For FathomVerse v0, the authors included all players with an F1 \textgreater= 0.8, regardless of total annotations.}
        \label{fig:player_threshold_num_ann}
    \end{subfigure}
    \caption{Player performance in FathomVerse v0 dataset. (a) Precision (blue line) and recall (organge line) of the dataset as a function of F1 threshold of players. Player's reaching an F1 threshold of 0.8 are indicated by black \textit{x}. (b) Precision and recall of data from players achieving F1 \textgreater= 0.8 as a function of total number of annotations.}
    \label{fig:player_threshold}
\end{figure}

In order to achieve this, the FathomVerse v0 dataset was constrained by thresholding participating players by their F1 score, which was calculated by comparing their annotations to expertly labeled images that were generated prior to the experiment. The precision and recall of the dataset varied when only players above a given F1 score were considered (Fig.~\ref{fig:player_threshold_f1}. Precision and recall are calculated by comparing the player consensus label to our expert labels. Consensus labels are determined by counting binary player votes on each \textit{bounding box-animal} combination queried in the game. Only \textit{bounding boxes-animal} tuples that received at least three votes are included in the consensus dataset. The precision and recall trends increase markedly when the players are required to have an F1 greater than 0.7 (Fig.~\ref{fig:player_threshold_f1}b). Increasing the F1 threshold to 0.95, for example, would have eliminated most player annotations, thereby making the dataset less useful. Annotations from players with an F1 score of \textgreater= 0.8 were included in the dataset. In total, 335 out of 863 game players contributed to the FathomVerse v0 dataset. 

Thresholding by number of contributed annotations was also considered, since players who contributed more annotations might be more trustworthy (Fig.~\ref{fig:player_threshold_num_ann}). However, thresholding players by their total contribution would not have increased the consensus dataset recall, and thus number of annotations was not used as a criteria for selecting contributing players to for the dataset.

\begin{figure}[h!]
    \centering
    \includegraphics[width=0.9\linewidth]{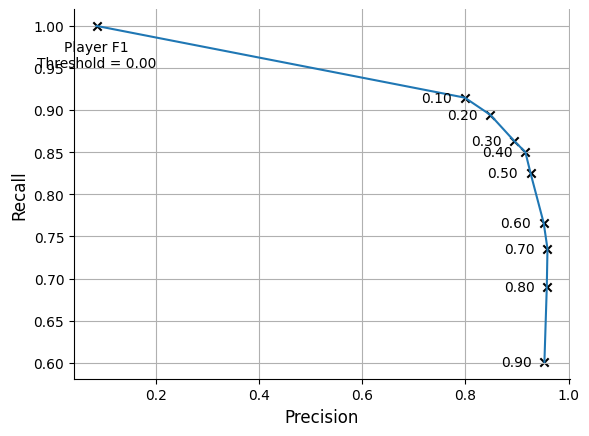}
    \caption{Precision and Recall curve showing the quality of the player consensus dataset, FathomVerse v0, constructed using all player annotations from players with an F1 \textgreater= 0.8, and at least 3 annotations per (bounding box, animal) tupple.}
    \label{fig:fathomversev0-pr}
\end{figure}

Figure \ref{fig:fathomversev0-pr} shows how FathomVerse v0 compares to a marine biologist oracle. An expert marine biologist on the author team reviewed 2084 of the 3843 FathomVerse v0 images that were previously unlabeled. For our detector training experiments, we chose to count normalized consensus labels of \textless 0.5 as negative labels and \textgreater= 0.5 as positive labels. According to Fig. \ref{fig:fathomversev0-pr}, a binarized version of FathomVerse v0 at that threshold has a precision of 0.93 and recall of 0.83. We believe this dataset should still work well to train detectors because of the well-known robustness of deep networks to label noise\cite{rolnick2017deep,krause2016unreasonable}, and we verify our hypothesis in the following section.

\begin{figure}[h!]
    \centering
    \includegraphics[width=0.9\linewidth]{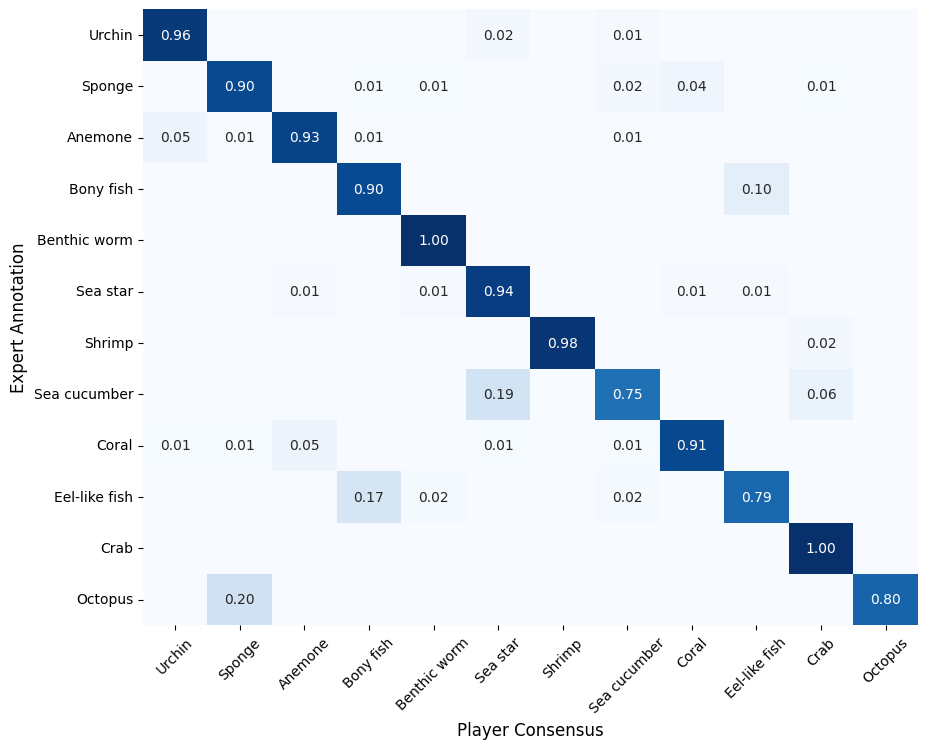}
    \caption{Normalized confusion matrix for FathomVerse v0 compared to expert annotations.}
    \label{fig:fathomversev0-confusion}
\end{figure}

\begin{figure}
    \centering
    \includegraphics[width=0.3\linewidth]{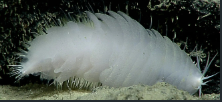}
    \includegraphics[width=0.3\linewidth]{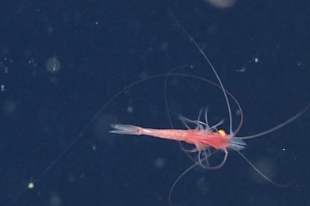}
    \includegraphics[width=0.3\linewidth]{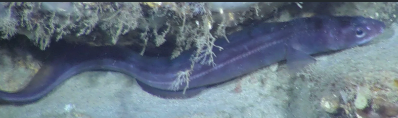}
    \includegraphics[width=0.3\linewidth]{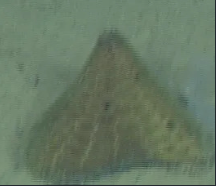}
    \includegraphics[width=0.3\linewidth]{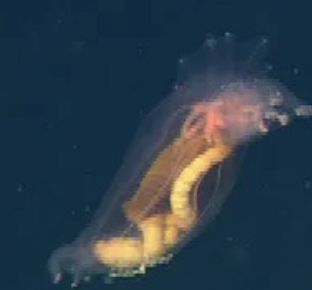}
    \includegraphics[width=0.3\linewidth]{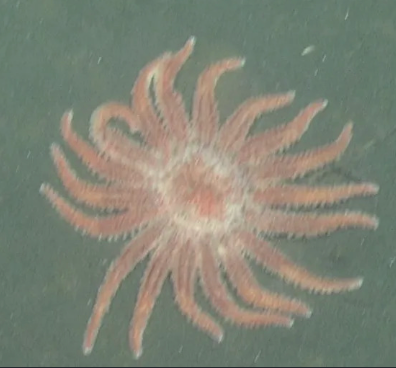}
    \includegraphics[width=0.3\linewidth]{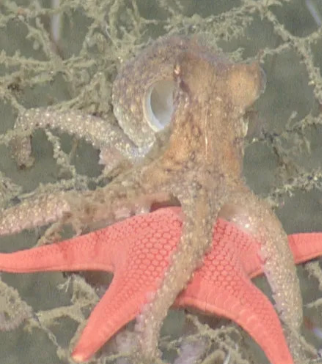}
    \includegraphics[width=0.3\linewidth]{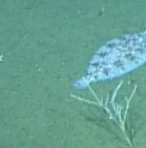}
    \includegraphics[width=0.3\linewidth]{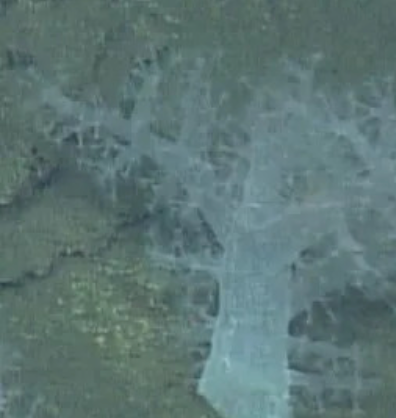}
    \caption{Example regions of interest from FathomVerse v0. The animals in order from the top-left are \textit{benthic worm}, \textit{shrimp}, \textit{eel-like fish}, \textit{urchin}, \textit{sea cucumber}, \textit{sea star}, \textit{octopus} and \textit{sea star}, \textit{bony fish} and \textit{coral}, and \textit{coral}. Notice the challenges to detection such as the body shapes of the \textit{benthic worm}, \textit{shrimp}, and \textit{eel-like fish} in the top row, the background of the \textit{urchin} and \textit{coral}, and the transparent body of the \textit{sea cucumber}.}
    \label{fig:dataset-examples}
\end{figure}

A confusion matrix of the FathomVerse v0 dataset (Fig. \ref{fig:fathomversev0-confusion}) and sample representative animal imagery (Fig.~\ref{fig:dataset-examples}) presents several shortcomings and challenges of the dataset. \textit{Octopus} was confused more often than expected, and on detailed review of those annotations, this was because \textit{octopus} were often confused by vampire squid despite being listed as a confounding species in the \textit{octopus} mission briefing. Less obvious player mistakes include the confusion of \textit{acorn worms} for \textit{benthic worms} since players often erroneously concluded that \textit{benthic worms} were inclusive of \textit{acorn worms}. Future versions of the game will include mission briefings on \textit{acorn worms}. One large source of player confusion was how to label a bounding box that contained two animals, which occurs frequently. This explains why \textit{octopus} and \textit{sponge} were confused; there were differing opinions between the players and experts as to which animal is the dominant animal in a given region of interest. 

\section{FathomVerse Detectors}% - GEN}
\label{sec:detectors}
Besides helping scientists record the presence and population of ocean animals in diverse global locations, the purpose of the FathomVerse v0 dataset is to train animal detectors that can be used to accelerate processing of visual data by researchers. The variable and distinct visual conditions of the deep sea may also provide challenges that advance detector architecture research in the future. Here we present two baseline detector models and highlight where a commonly used model type is failing at the task of fine-grained detection.

\begin{figure}[h]
    \centering
    \includegraphics[width=\linewidth]{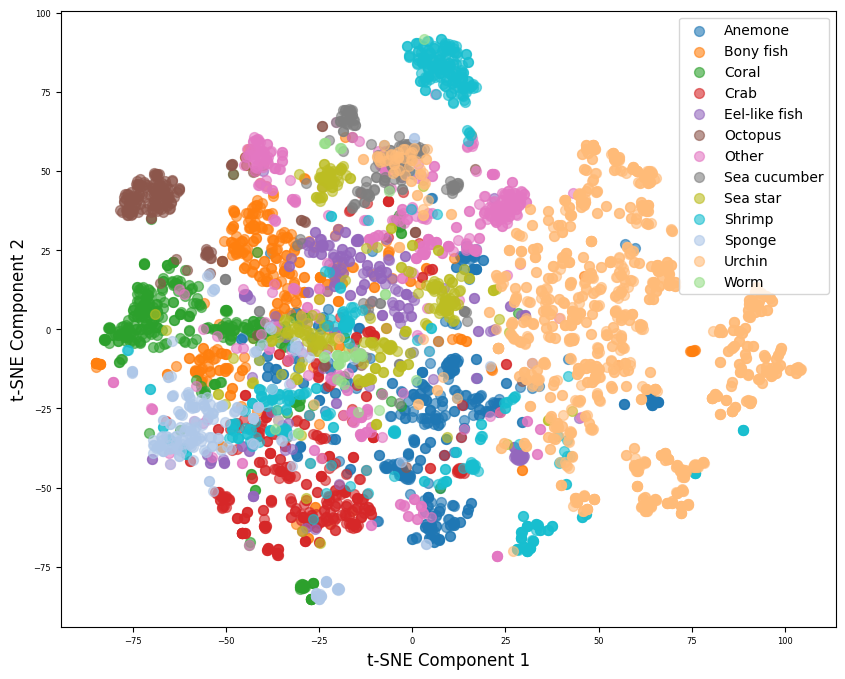}
    \caption{t-SNE plot showing clustering of the 12 different animal mission groups found in the FathomVerse v0 dataset.}
    \label{fig:tsne}
\end{figure}

We began by reassuring ourselves that the animal concept groups were relatively separable in an appropriate embedding space (Fig.~\ref{fig:tsne}). The t-SNE\cite{van2008visualizing} plot, was constructed using SPPF\cite{he2015spatial} feature vector for the center of each ROI in the FathomVerse v0 dataset, which was extracted via an adapted YOLOv8\cite{redmon2016you,varghese2024yolov8} architecture and trained on an 80k train/10k validation image dataset from the FathomNet Database images captured by the authors' institution using the Ultralytics Python package\cite{ultralytics}. Even in 2D, samples from the 12 animal groups appear to cluster tightly. Overlapping clusters contain animals that are morphologically similar or appear in similar contexts. Due to the qualitative clustering, we expected transfer learning from the location where FathomNet Database training data were generated (Monterey Bay) to the Octopus Garden to perform relatively well, and to Musicians Seamounts to perform poorly given their different geolocation\cite{hsu2017learning}.

\begin{figure}
    \centering
    \begin{subfigure}[t]{0.5\linewidth}
        \centering
        \includegraphics[width=\linewidth]{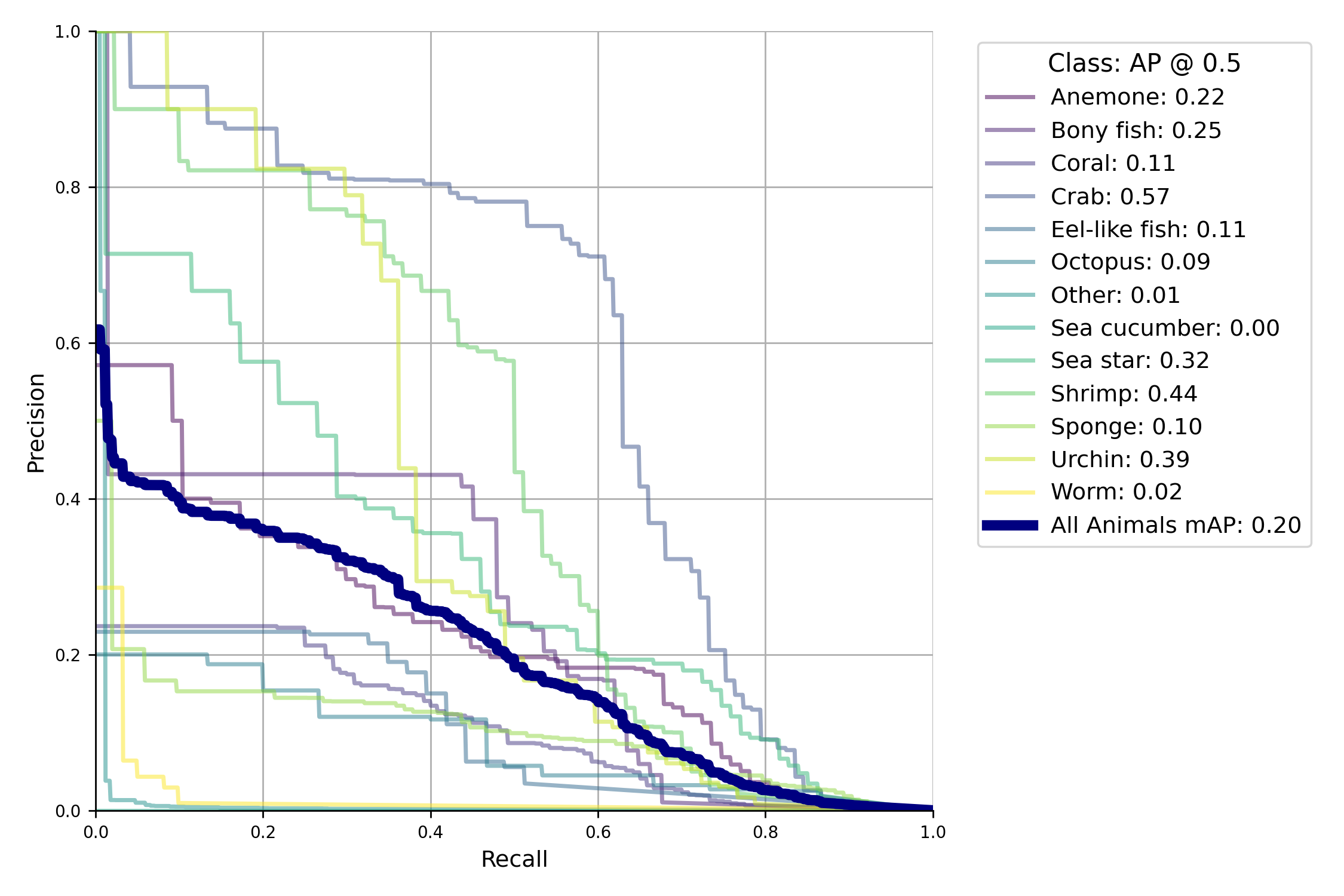}
        \caption{ }
    \end{subfigure}%
    ~
    \begin{subfigure}[t]{0.5\linewidth}
        \centering
        \includegraphics[width=\linewidth]{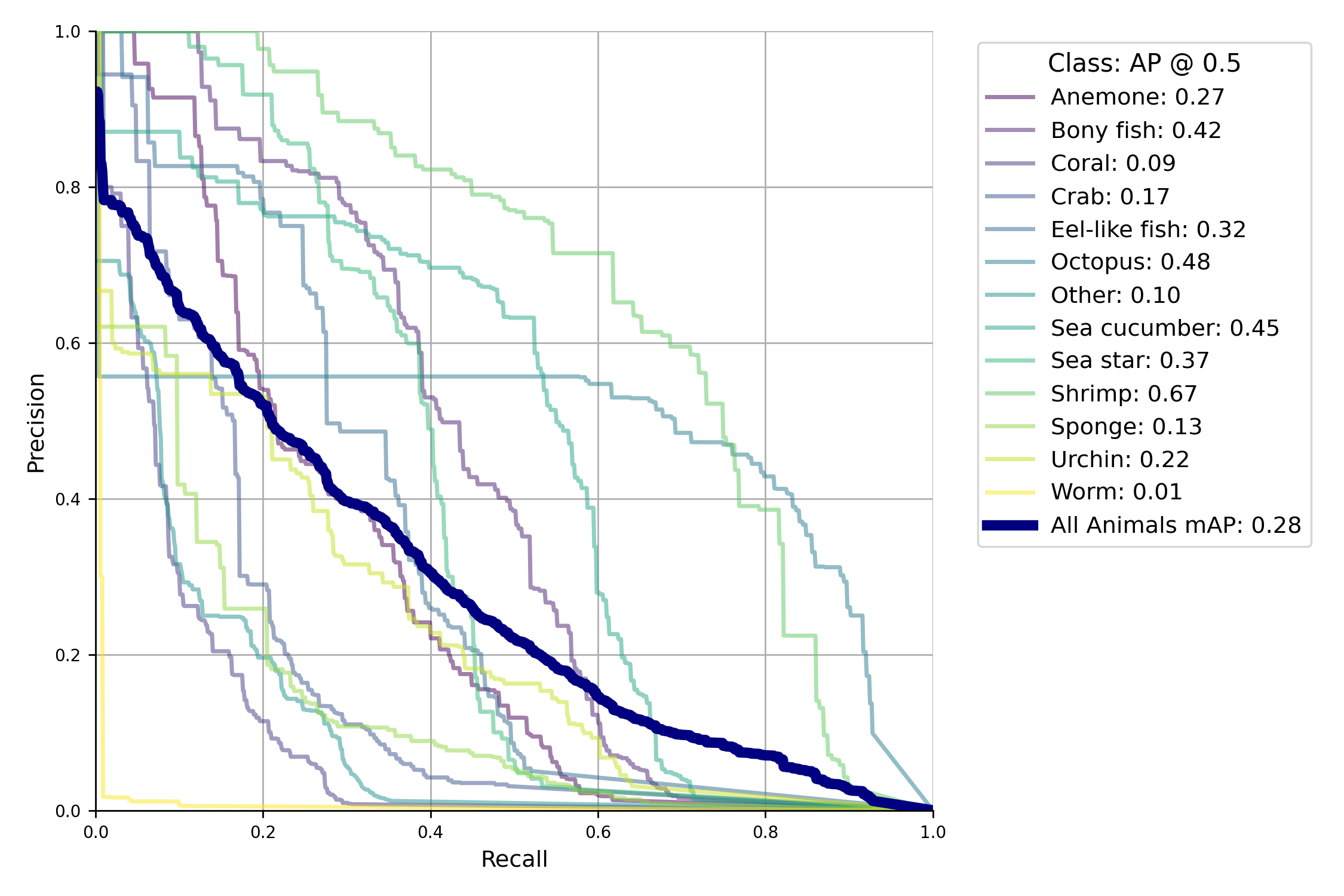}
        \caption{ }
    \end{subfigure}%
    \caption{Precision-recall (PR) curves for detections at IoU 0.5 from the FathomNet Detector. (a) PR Curve for the 13-class FathomNet YOLOv8 Detector, evaluated on FathomVerse v0 samples from the Octopus Garden. (b) PR Curve for the 13-class FathomNet YOLOv8 Detector, evaluated on FathomVerse v0 samples from the Musicians Seamounts.}
    \label{fig:FN-full-FV-test}
\end{figure}

\begin{figure}
    \centering
        \begin{subfigure}[t]{0.48\linewidth}
        \centering
    \includegraphics[width=1.0\linewidth]{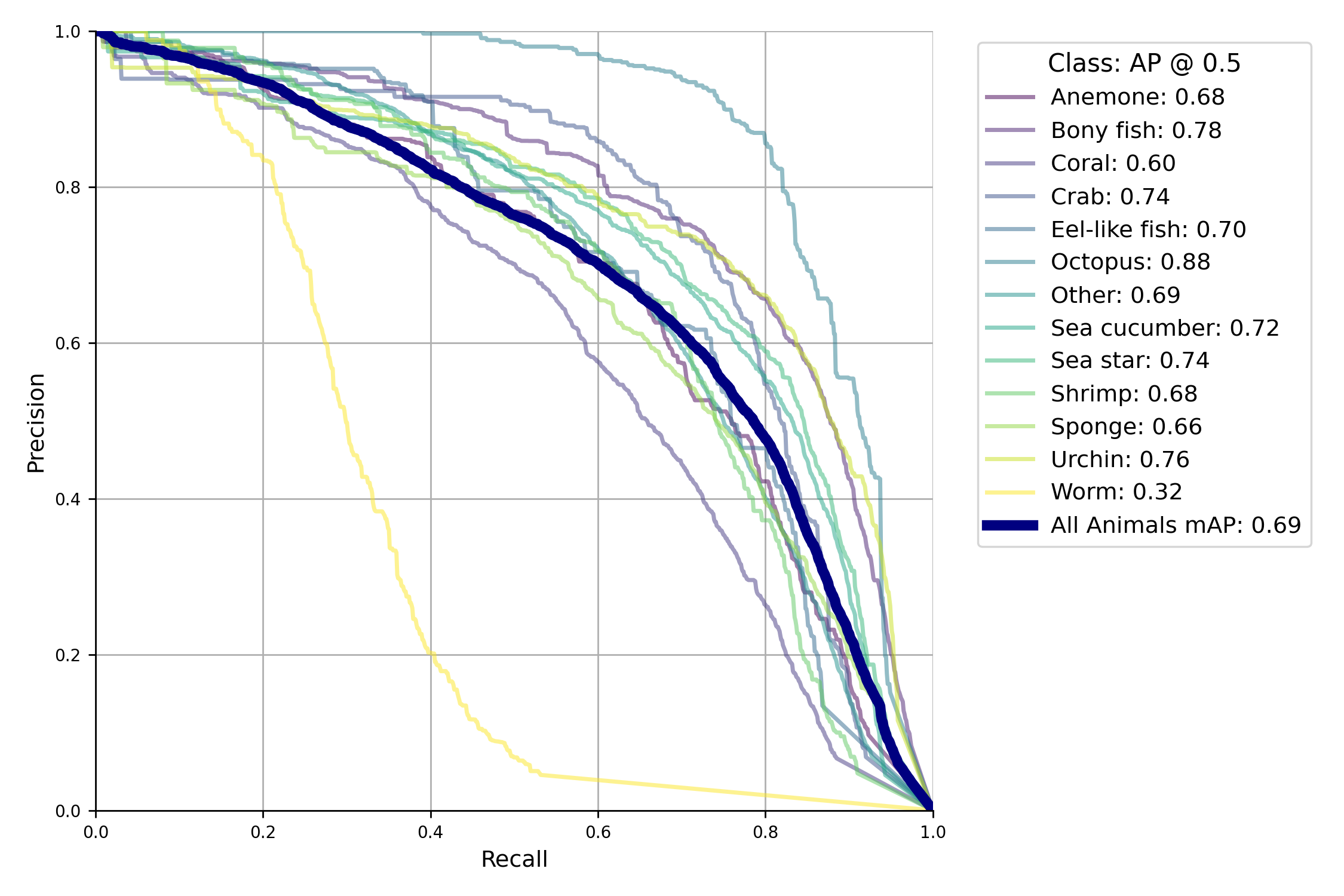}
    \caption{ }
    \label{fig:FN-FN-test}
    \end{subfigure}
    ~
        \begin{subfigure}[t]{0.48\linewidth}
        \centering
    \includegraphics[width=1.0\linewidth]{figures/eval_MBARI_only_FathomNet_model-model_on_Octopus_Garden_testset/PR_curve.png}
    \caption{ }
    \label{fig:FN-FV-test}
    \end{subfigure}
    \\
    \begin{subfigure}[t]{0.48\linewidth}
        \centering
    \includegraphics[width=1.0\linewidth]{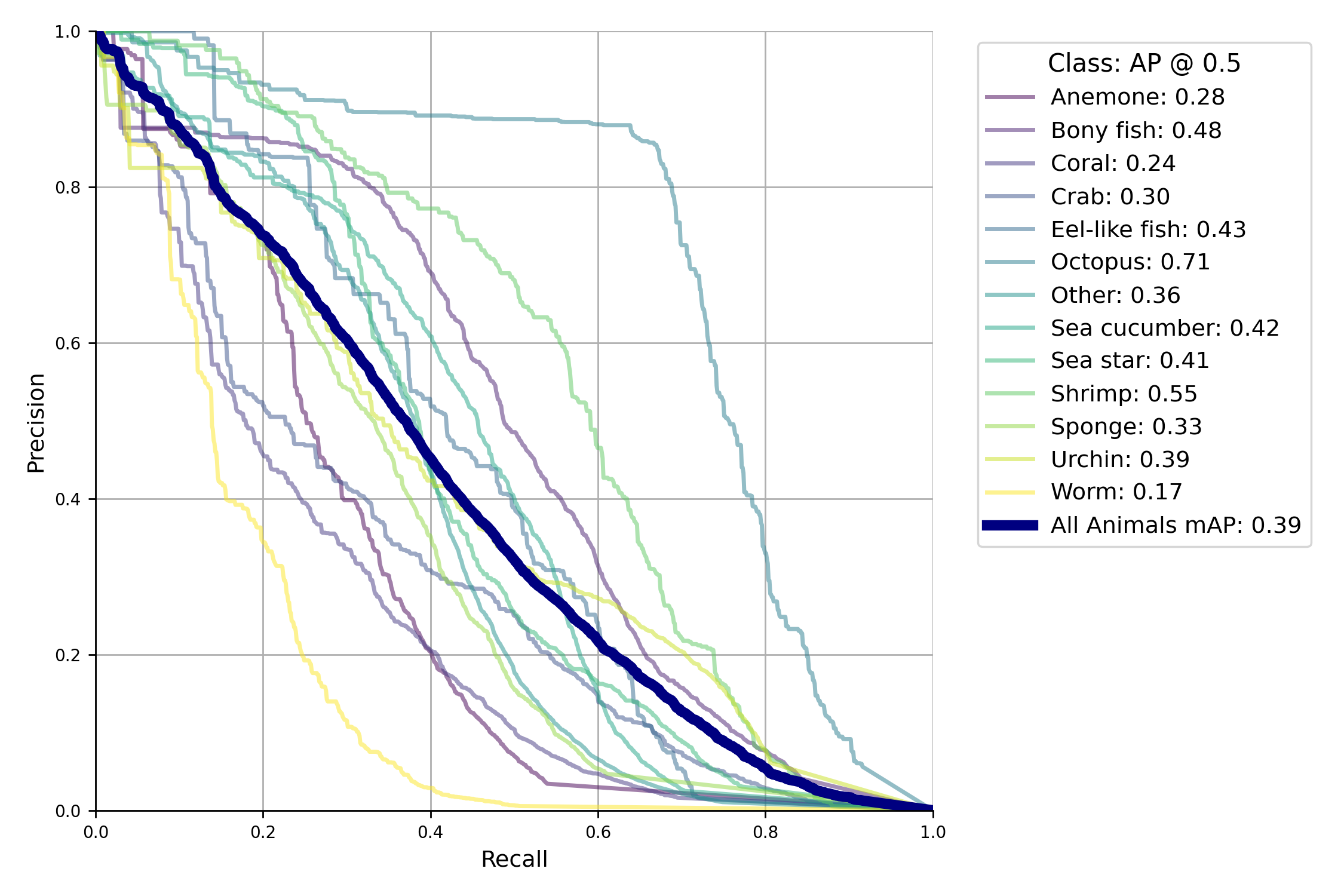}
    \caption{ }
    \label{fig:FV-FN-test}
    \end{subfigure}
    ~
    \begin{subfigure}[t]{0.48\linewidth}
        \centering
    \includegraphics[width=1.0\linewidth]{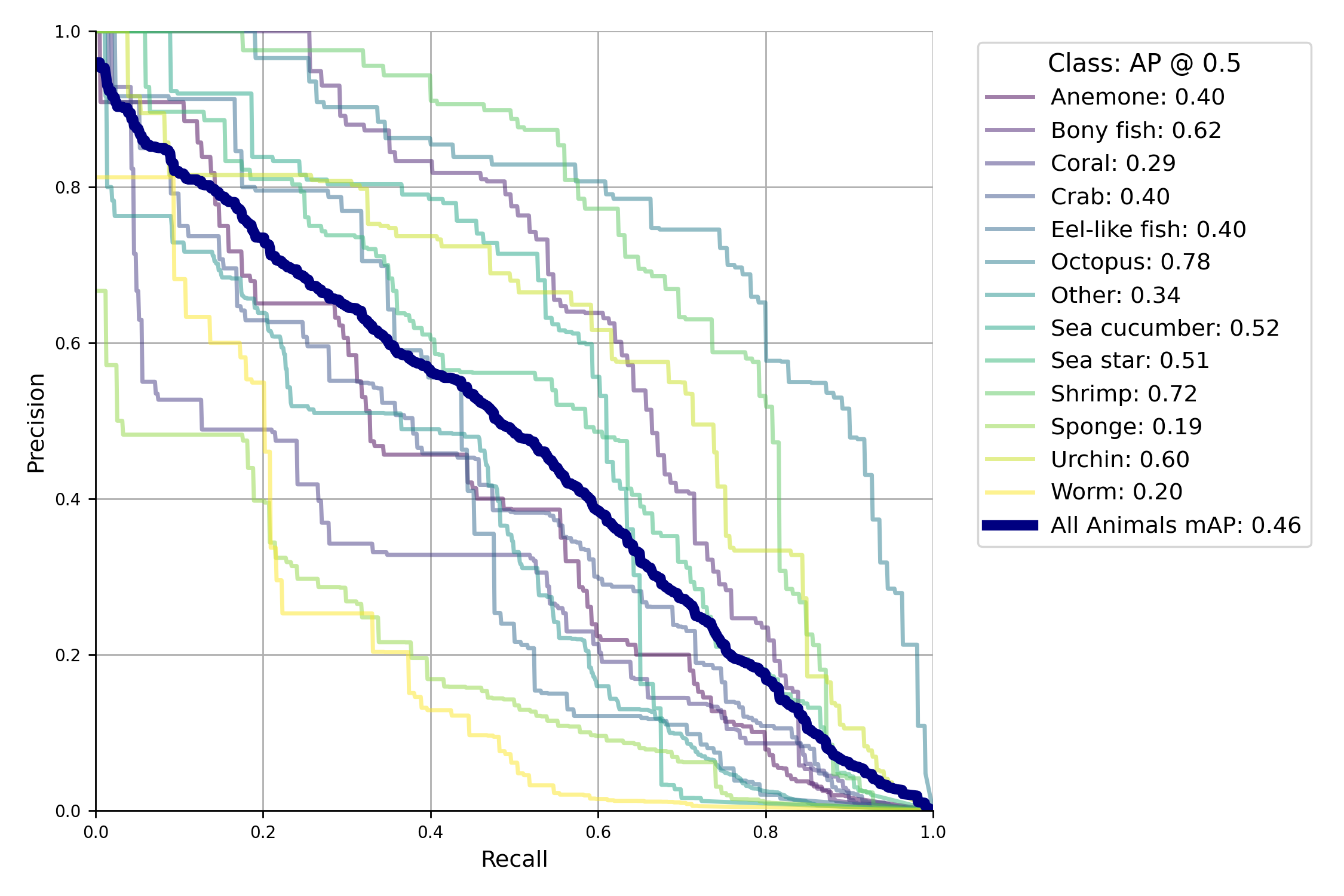}
    \caption{ }
    \label{fig:FV-FV-test}
    \end{subfigure}
    \caption{Precision-recall (PR) curves for detections at IoU 0.5 from the FathomNet and FathomVerse Detectors on the FathomNet and FathomVerse test sets. (a) Precision and recall (PR) curves for the FathomNet Detector evaluated on the 13-class FathomNet test set (10k images). (b) PR Curve for FathomNet Detector tested on the FathomVerse v0 test set (approximately 3k images from both new benthic locations). (c) PR Curve for FathomVerse Detector evaluated on the 13-class FathomNet test set. (d) PR Curve for FathomVerse Detector tested on the FathomVerse v0 test set.}
\end{figure}

\begin{figure}
    \centering
    \includegraphics[width=0.9\linewidth]{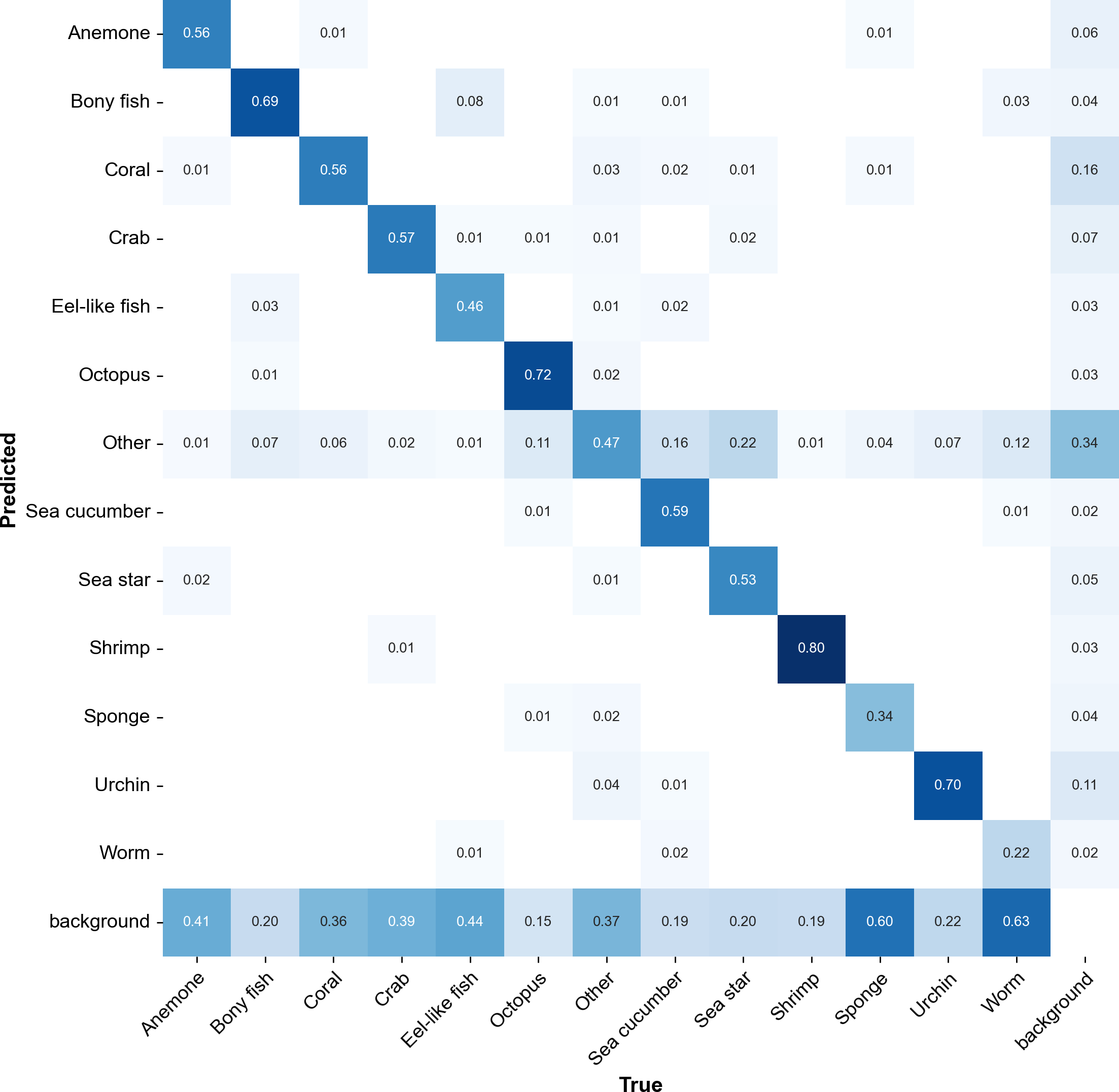}
    \caption{Normalized confusion matrix for the FathomVerse Detector tested on the FathomVerse v0 test set.}
    \label{fig:detector-confustion}
\end{figure}

Our first detector was trained using images made by the authors' institution and contributed to the open source dataset effort FathomNet Database\cite{katija2022fathomnet}. This dataset was split into test/val/training sets of approximately 10k/10k/80k images respectively. The FathomNet classes for each sample were remapped from their genus or species level annotations to the 12 classes in FathomVerse plus an \textit{Other} category for animals that didn't fit into the 12-class taxonomy. We selected the YOLO architecture, popular for fine-grained animal recognition in the wild\cite{xu2023fine}, for our baseline. Our baseline YOLOv8\cite{varghese2024yolov8} detector was trained with an Adam learning rate optimizer, initial learning rate of 0.001, for 25 epochs. The epoch limitation was selected as an early-stopping avoidance of overfitting to the training set, as observed on the FathomNet validation and test sets (Fig.~\ref{fig:FN-full-FV-test}).

The FathomNet 13-category detector performed relatively well (Fig. \ref{fig:FN-FN-test} on a large test dataset with a variety of animals belonging to the 13 groups, although the FathomNet train/test sets are sourced predominantly from the northeast Pacific. The FathomNet detector performed poorly on images from the Musicians Seamount, a new location, and also on the Octopus Garden. The Octopus Garden is a warm water spring found in the deep sea, and is an octopus brooding hotspot dominated by just a few unique species. Although found in Monterey Bay, which is where most of the FathomNet Dataset has been collected, the Octopus Garden is quite unique and poor performance at this location was not unlikely. Cosidering the drop in performance from the FathomNet test set (Fig. \ref{fig:FN-FN-test}) to the FathomVerse test set (Fig. \ref{fig:FN-FV-test}; even split of Musicians Seamount and Octopus Garden images), we would expect that fine-tuning the FathomNet detector on a FathomVerse v0 training set would improve performance.

Our second detector, FathomVerse Detector, was trained using a test/val/train split of 40\%/10\%/50\% respectively of the FathomVerse v0 dataset. We reserved a larger share of the dataset for testing in order to properly evaluate the rarer animal classes like \textit{Octopus}. We updated the FathomNet Detector by freezing the nine-layer SPPF backbone and allowing the classification and bounding box proposal task heads to update. We ran training for only 5 epochs due to the size of the training set (to prevent overfitting as observed on the validation set only), with an initial learning rate of 5e-6.

\begin{figure*}[h]
    \centering
            \begin{subfigure}{0.5\linewidth}
        \centering
    \includegraphics[width=1.0\linewidth]{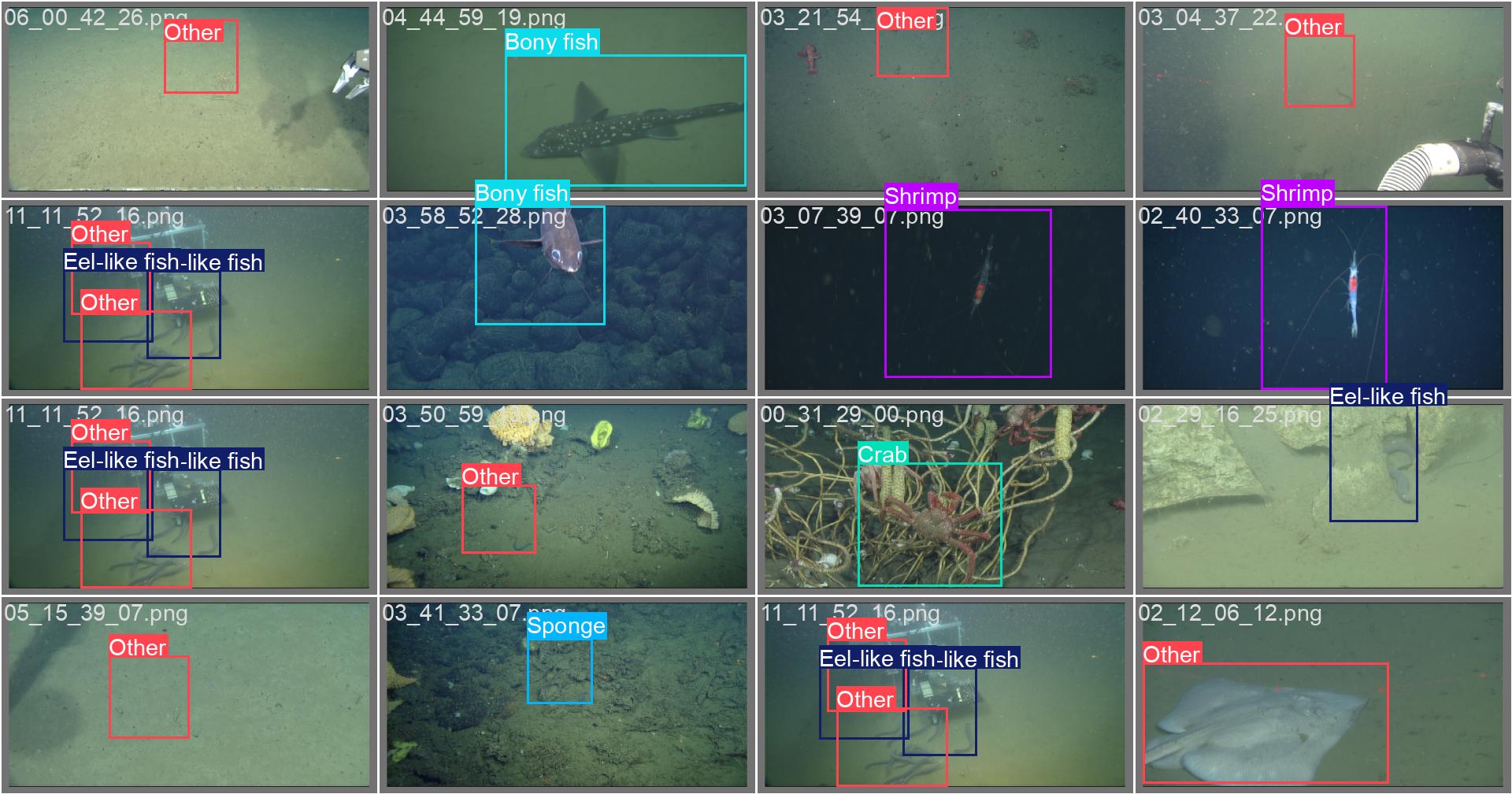}
            \caption{Labeled Test Samples}
            \label{fig:eel-like-labels}
    \end{subfigure}%
    ~
            \begin{subfigure}{0.5\linewidth}
        \centering
    \includegraphics[width=1.0\linewidth]{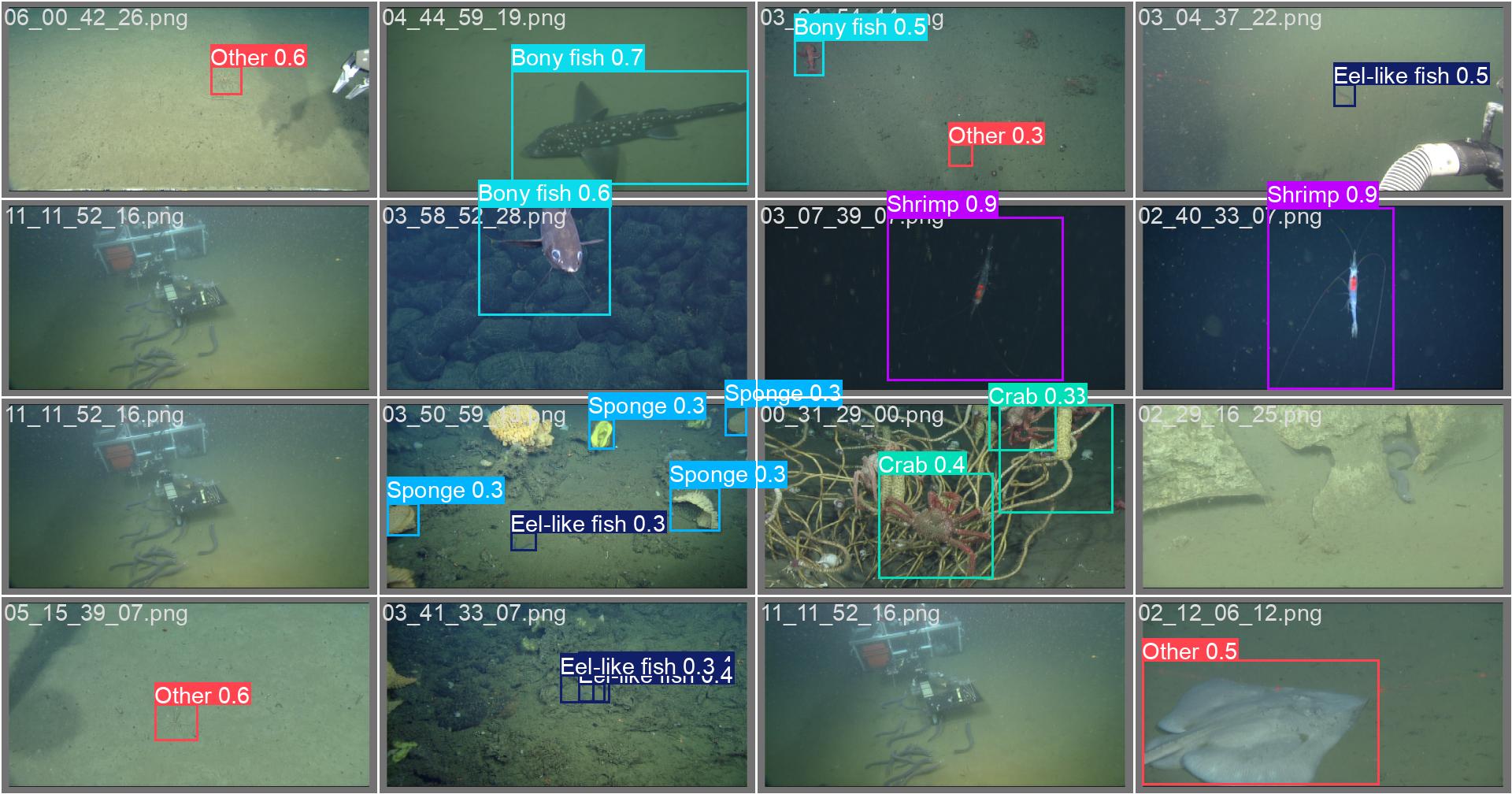}
                \caption{Detector Predictions}
                \label{fig:eel-like-detections}
    \end{subfigure}
    \\
            \begin{subfigure}{0.5\linewidth}
        \centering
        \includegraphics[width=1.0\linewidth]{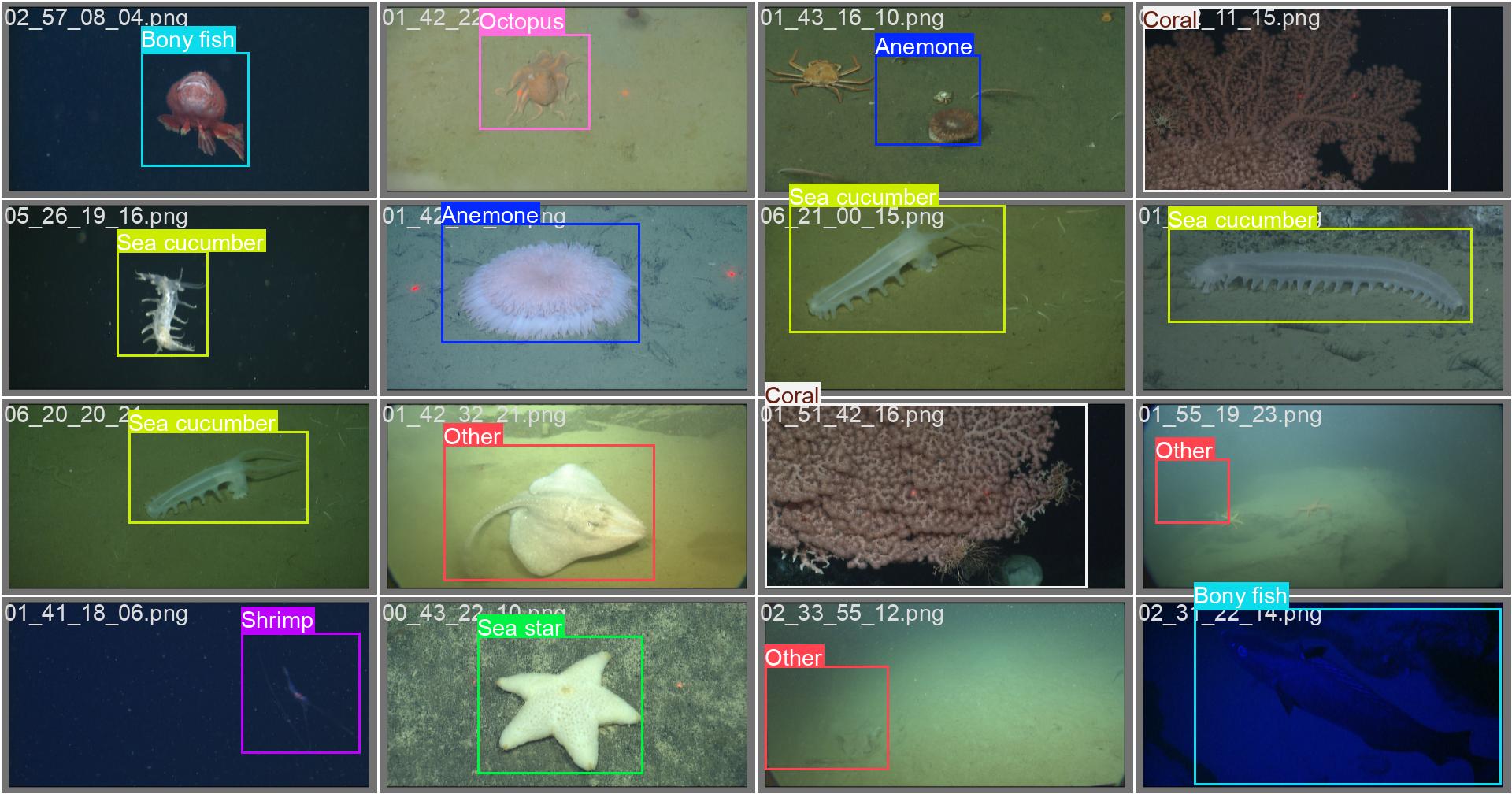}
                    \caption{Labeled Test Samples}
                    \label{fig:worm-labels}
    \end{subfigure}%
    ~
                \begin{subfigure}{0.5\linewidth}
        \centering
    \includegraphics[width=1.0\linewidth]{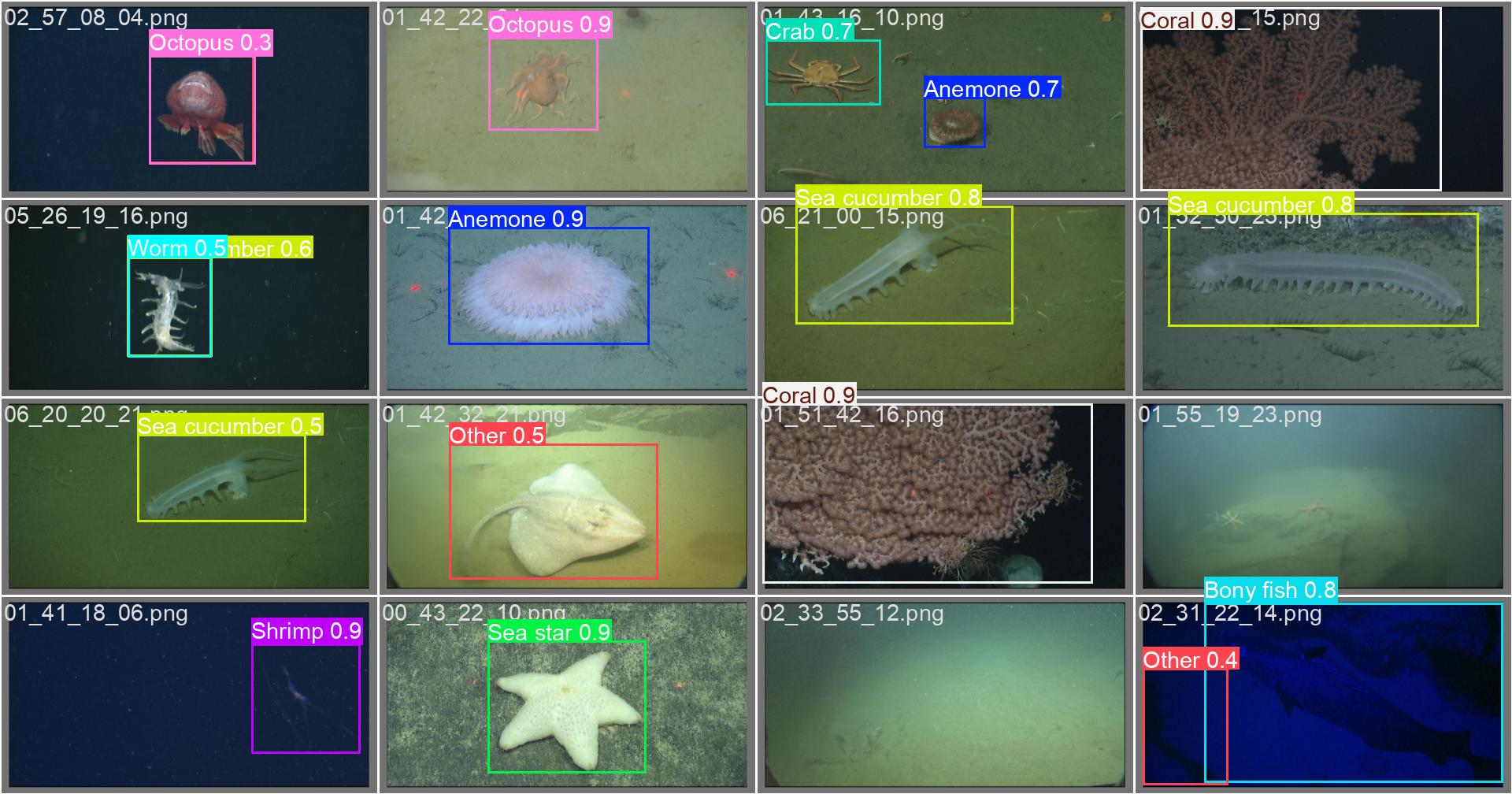}
                \caption{Detector Predictions}
                \label{fig:worm-detections}
    \end{subfigure}%
    \\
            \begin{subfigure}{0.5\linewidth}
        \centering
        \includegraphics[width=1.0\linewidth]{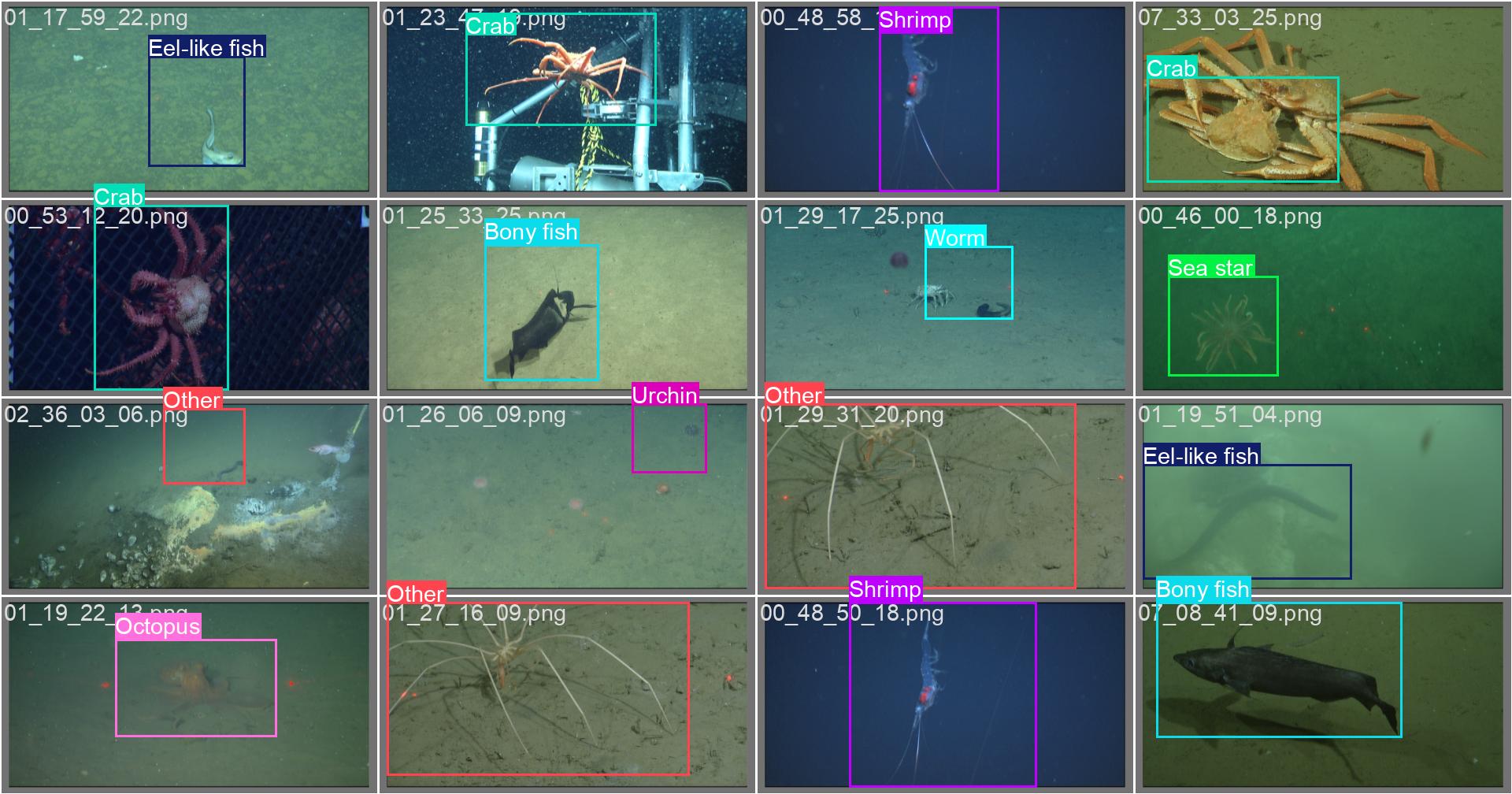}
                    \caption{Labeled Test Samples}
                    \label{fig:crab-labels}
    \end{subfigure}%
    ~
                \begin{subfigure}{0.45\linewidth}
        \centering
    \includegraphics[width=1.0\linewidth]{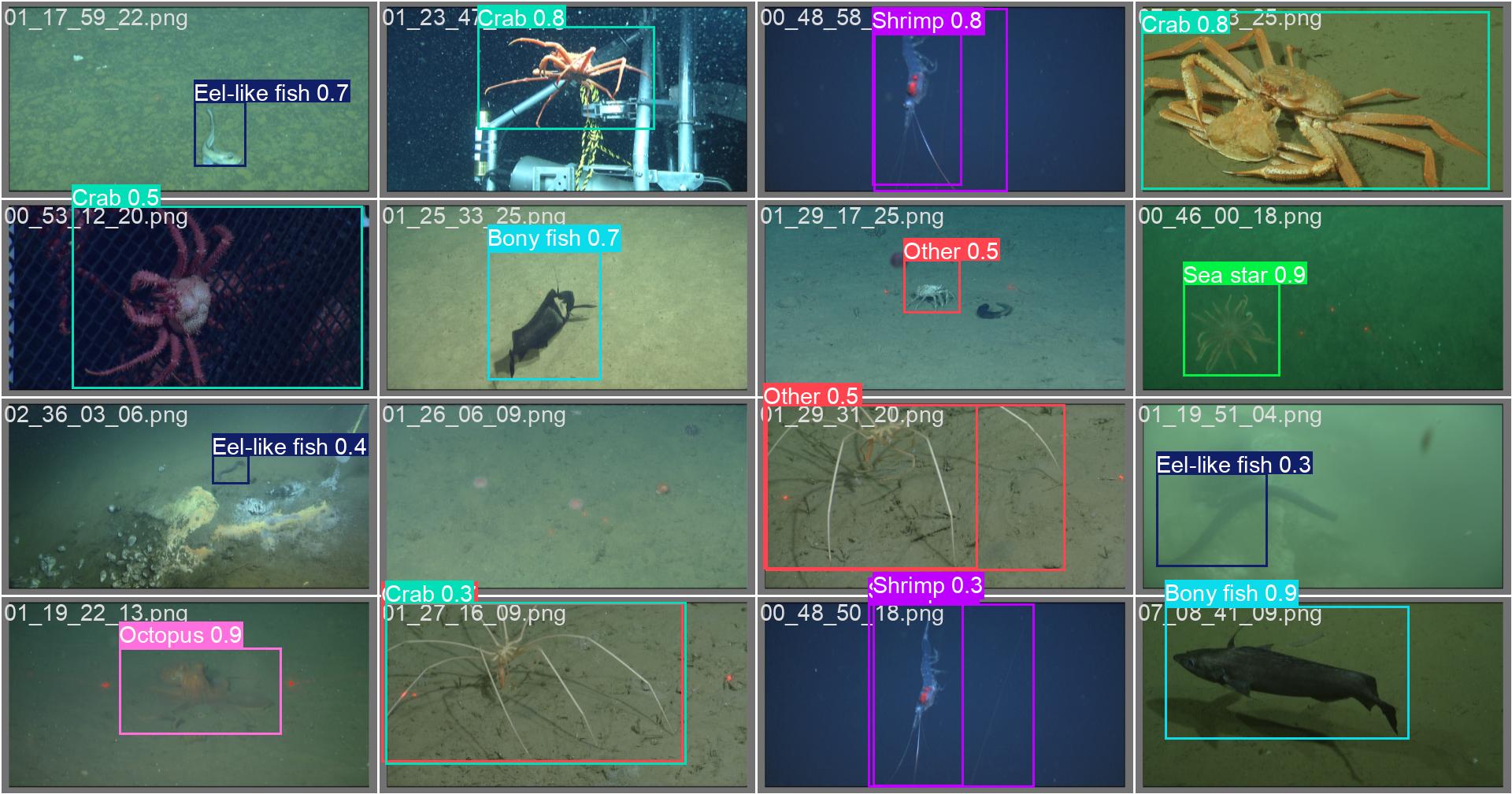}
                \caption{Detector Predictions}
                \label{fig:crab-detections}
    \end{subfigure}%
    \caption{Anecdotal results for the YOLOv8 FathomVerse Detector tested on the FathomVerse v0 test set. Note the animals, especially Eel-like Fish, which are confused for background. In the last row there is an example of the detector confusing a Sea Spider for a Crab, which was an educational example of a confounding category for the game players!}
    \label{fig:test-detections}
\end{figure*}

Figures \ref{fig:FV-FN-test} and \ref{fig:FV-FV-test} show how the FathomVerse Detector behaves on FathomNet data (mostly from Monterey Bay) and FathomVerse data (Musicians Seamounts and the Octopus Garden). Modest fine-tuning helps, but much more data is needed for generalization of these animal concepts. The modest performance of the FathomVerse Detector on both the FathomNet and FathomVerse test sets, even after training on more than 80,000 images, suggests that innovation in detector architecture or training pipeline may be necessary to cope with the variability of appearance and location of ocean animals.

Investigating the details of FathomVerse detection failures, we see from the confusion matrix (Fig.~\ref{fig:detector-confustion}) that animals are frequently confused with background material. \textit{Worms}, which burrow into substrate dirt and mud, are especially easy to miss. Figure~\ref{fig:test-detections} shows anecdotal successes and failures of the detectors. Figures \ref{fig:eel-like-labels} and \ref{fig:eel-like-detections} show how the FathomVerse detector confuses \textit{Eel-like fish} with the background mud, similar to the \textit{Worm} misses noted in Fig. \ref{fig:detector-confustion}. Figures \ref{fig:worm-labels} and \ref{fig:worm-detections} show how \textit{Sea Cucumbers} can be confused with \textit{Worms}. This is an easy mistake for an untrained player, but we would hope for more from a detector that should approach the ability of an expert taxonomist. Figures~\ref{fig:crab-labels} and \ref{fig:crab-detections} show the complicated and unusual-in-land-animal interactions from animals like crabs, and shows an example occlusion by members of the same class.

\section{Conclusion and Future Work}% - Kakani}
\label{sec:conclusion}
The FathomVerse v0 dataset achieves high precision and recall through player consensus, even for non-expert non-enthusiast, general public audiences, demonstrating the value of gamifying annotation. Future iterations of this dataset will improve from wider selection of animal categories and background substrate labeling. We hope this challenging dataset will push the envelope for transfer learning, novel category discovery, and conservation science.

%%%%%%%%% REFERENCES
{\small
\bibliographystyle{ieee_fullname}
\bibliography{egbib}
}

\end{document}